%% file: neurips_2025.tex
\documentclass{article}

\PassOptionsToPackage{numbers, compress}{natbib}

\usepackage[preprint]{neurips_2025}

\usepackage[utf8]{inputenc} %
\usepackage[T1]{fontenc}    %
\usepackage{hyperref}       %
\usepackage{url}            %
\usepackage{booktabs}       %
\usepackage{amsfonts}       %
\usepackage{nicefrac}       %
\usepackage{microtype}      %
\usepackage[table]{xcolor}

\usepackage{tabulary}
\usepackage{etoolbox}

\newcommand{\highlightgreen}[1]{\colorbox[HTML]{d3ff9f}{\textbf{#1}}}

\newcommand{\highlightblue}[1]{\colorbox[HTML]{bae6fb}{\textbf{#1}}}

\definecolor{mycolor_blue}{HTML}{E7EFFA}
\definecolor{mycolor_green}{HTML}{E6F8E0}
\definecolor{mycolor_gray}{HTML}{ECECEC}
\definecolor{pearDark}{HTML}{2980B9}
\usepackage{pgfplots}
\usepackage{pgfplotstable}
\usepackage{sansmath}
\usepackage{helvet}
\usepackage{graphicx}
\usepackage{amsmath}
\usepackage{float}
\usepackage{tcolorbox}
\usepackage{etoc}
\usepackage{scrextend}
\usepackage{wrapfig}
\usepackage{titletoc}
\usepackage{subfigure}
\usepackage{multirow}

\newlength\mystoreparindent

\title{
RARE: Retrieval-Augmented Reasoning Modeling
}

\pgfplotsset{compat=1.18}
\author{
    \textbf{Zhengren Wang}$^\dagger$\textsuperscript{1,5},
    \textbf{Jiayang Yu}$^\dagger$\textsuperscript{3},
    \textbf{Dongsheng Ma}$^\dagger$\textsuperscript{4}\\
    \textbf{Zhe Chen}\textsuperscript{\textbf{2,8}}, 
    \textbf{Yu Wang}\textsuperscript{\textbf{2,8}},
    \textbf{Zhiyu Li}\textsuperscript{*\textbf{5,7}},
    \textbf{Feiyu Xiong}\textsuperscript{\textbf{5,7}} \\
    \textbf{Yanfeng Wang}\textsuperscript{\textbf{2,8}}, 
    \textbf{Weinan E}\textsuperscript{\textbf{1,2,5}},
    \textbf{Linpeng Tang}\textsuperscript{\textbf{5,6}},
    \textbf{Wentao Zhang}\textsuperscript{*\textbf{1,5,6}}\\
    \textsuperscript{1}Peking University 
    \textsuperscript{2}Shanghai Jiao Tong University 
    \textsuperscript{3}Northeastern University 
    \textsuperscript{4}Nankai University \\
    \textsuperscript{5}Institute for Advanced Algorithms Research, Shanghai 
    \textsuperscript{6}OriginHub Tech.
    \textsuperscript{7}MemTensor Tech.\\
    \textsuperscript{8}Shanghai Artificial Intelligence Laboratory \\
    \texttt{wzr@stu.pku.edu.cn}, 
    \texttt{\{lizy,xiongfy\}@iaar.ac.cn}, 
    \texttt{wangyanfeng622@sjtu.edu.cn}\\
    \texttt{weinan@math.pku.edu.cn},
    \texttt{linpengt@originhub.tech}, 
    \texttt{wentao.zhang@pku.edu.cn}
    \vspace{-2em}
}

\begin{document}

\deffootnote[1.5em]{1.5em}{1em}{}
\renewcommand{\thefootnote}{\fnsymbol{footnote}}
\footnotetext{
$\dagger$ Equal contribution;  * Corresponding author. \\
}
\renewcommand{\thefootnote}{\arabic{footnote}}

\maketitle

\input{Sections/Abstract}

\input{Sections/Introduction}

\input{Sections/Related_Work}

\input{Sections/Method}

\input{Sections/Experiments}

\input{Sections/Conclusion}

{ %
\bibliographystyle{plainnat} \bibliography{neurips_2025} }

\input{Sections/Appendix}

\input{Sections/Check_List}

\end{document}

%% file: Sections/Abstract.tex
\begin{abstract}
\label{sec:abstract}
Domain-specific intelligence demands specialized knowledge and sophisticated reasoning for problem-solving, posing significant challenges for large language models (LLMs) that struggle with knowledge hallucination and inadequate reasoning capabilities under constrained parameter budgets. Inspired by Bloom's Taxonomy in educational theory, we propose Retrieval-Augmented Reasoning Modeling (RARE), a novel paradigm that decouples knowledge storage from reasoning optimization. RARE externalizes domain knowledge to retrievable sources and internalizes domain-specific reasoning patterns during training. Specifically, by injecting retrieved knowledge into training prompts with masked losses, RARE transforms learning objectives from rote memorization to contextualized reasoning. It enables models to bypass parameter-intensive memorization and prioritize the development of higher-order cognitive processes. Extensive experiments demonstrate that lightweight RARE-trained models (e.g., Llama-3.1-8B) could achieve state-of-the-art performance, surpassing retrieval-augmented GPT-4 and DeepSeek-R1 up to approximately 20\% accuracy. RARE establishes a paradigm shift where maintainable external knowledge bases synergize with compact, reasoning-optimized models, collectively driving more scalable domain-specific intelligence.

\end{abstract}

%% file: Sections/Introduction.tex
\begin{figure}[H]
     \centering
     \vspace{-1em}
     \includegraphics[width=\textwidth]{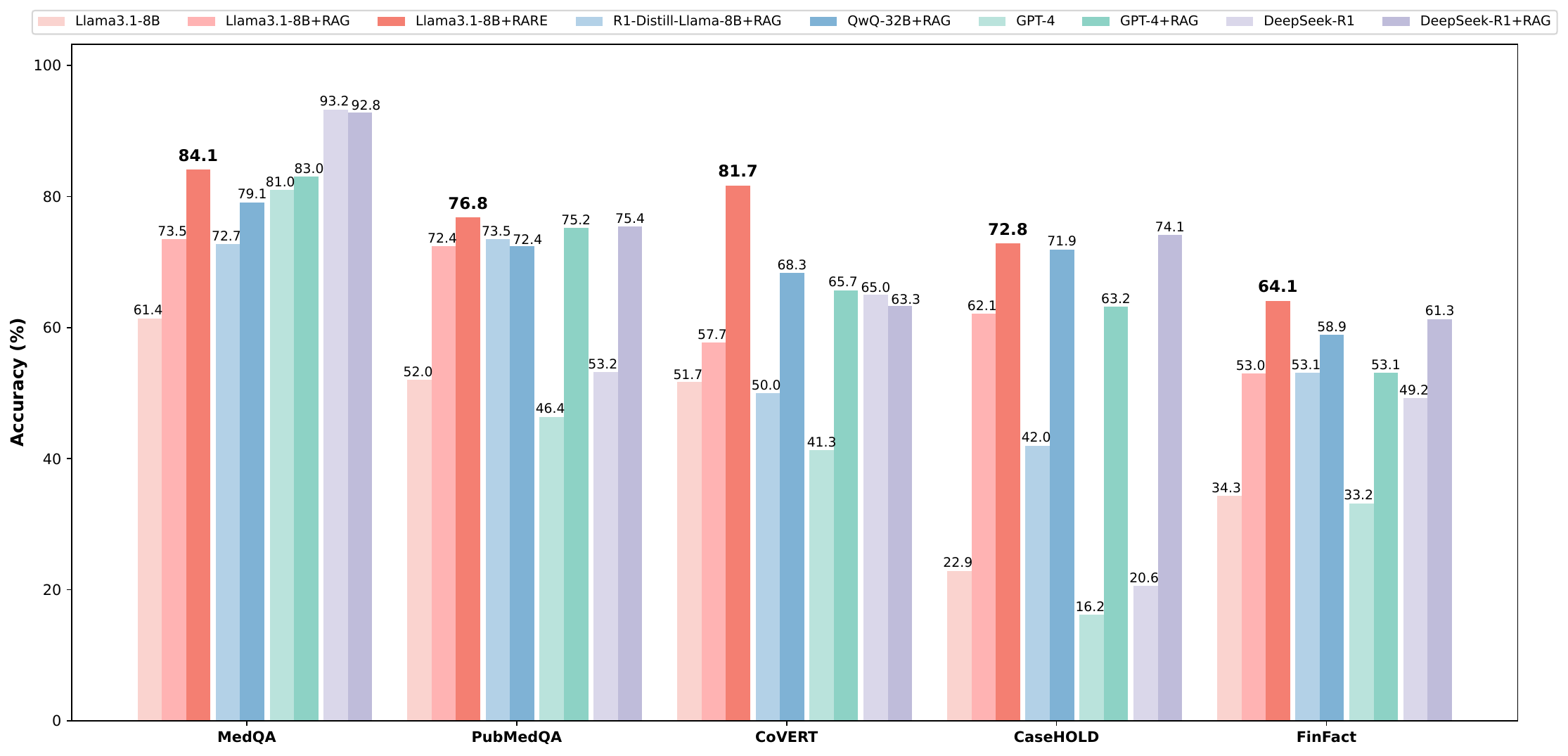}
     \vspace{-1.5em}
     \caption{ Performance of RARE versus baselines on benchmarks.}
    \label{fig:Benchmark}
\end{figure}

\section{Introduction}
\label{sec:introduction}
Large language models (LLMs), trained on vast corpora with billion-scale parameters, have demonstrated remarkable capabilities across diverse general-domain knowledge and reasoning tasks \citep{brown2020language, wei2022chain}. These models have revolutionized multiple application domains, such as mathematical reasoning \citep{wei2022chain, brown2020language} and task automation \citep{patil2023gorilla, yao2022react, wang2023voyager}.
However, there is an increasing need for \textit{domain-specific intelligence}, to tackle tasks involving specialized knowledge and reasoning capabilities. These tasks are prevalent in both specialized applications like medical specialist LLMs \citep{pham2024towards,pal2023med,morley2020ethics}, and general-purpose systems like open-domain generalist LLMs \citep{openai2023gpt4,guo2025deepseek} for diverse user scenarios.

\begin{figure}[htbp!]
     \centering
     \includegraphics[trim={0 0 0 0}, clip, scale=0.6]{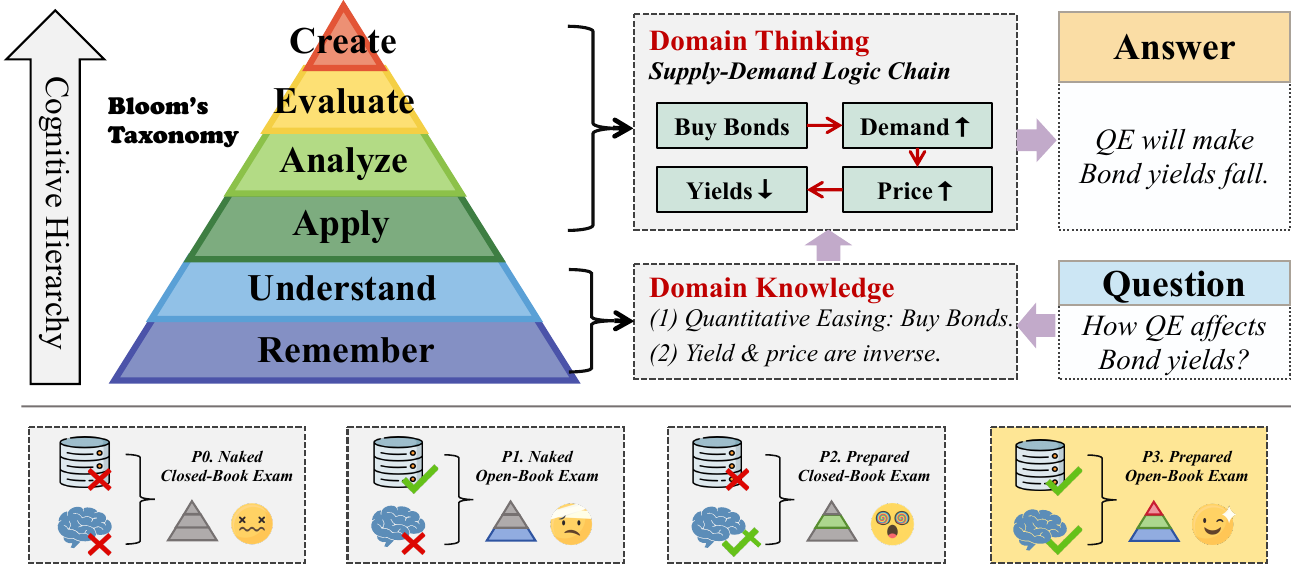}
     \caption{Motivation of RARE. Left: A pyramid-shaped Bloom’s Taxonomy, illustrating the cognitive hierarchy from basic "Remember" to advanced "Evaluate" and "Create" levels. Right: The correspondence between Domain Knowledge and Domain Thinking with Bloom’s cognitive hierarchy (example related to government bond yields). In contrast to domain knowledge, domain thinking corresponds to the higher-order cognitive process—although relatively \textit{\textbf{rare}}, it plays a crucial role.}
    \label{fig:Overview}
\end{figure}

As illustrated in Fig. \ref{fig:Overview}, the limitations and challenges in domain-specific contexts primarily stem from two key dimensions. \textit{\underline{Domain knowledge}}: Although billion-scale parameters counts to memorize, due to the parametric representation and long-tail nature of domain knowledge such as Power-law or Zipf's law distributions \citep{clauset2009power,newman2005power}, the hallucination phenomenon is still notoriously serious (LLMs as knowledge stores); \textit{\underline{Domain thinking}}: Beyond knowledge hallucination, vanilla LLMs also struggle with domain-specific reasoning, which requires the sophisticated application of both domain-specific knowledge and thinking skills (LLMs as reasoning engines). 
Both challenges highlight the critical research problem: \textit{the effective integration of domain-specific knowledge and reasoning capabilities, particularly under constrained parameter budgets.}

From the perspective of knowledge and reasoning acquisition, existing approaches can be intuitively categorized into three distinct paradigms:
\begin{itemize}
    \item[P0.] \textbf{Naked Closed-book Exam}: 
    Directly invoking general-purpose models without domain adaptation. These approaches suffer from poor performance due to the absence of both domain-specific knowledge and reasoning capabilities.
    \item[P1.] \textbf{Naked Open-book Exam}: 
    Retrieval-augmented generation (RAG) methods that address knowledge limitations through external information retrieval \citep{survey}. However, these approaches primarily focus on knowledge supplementation and neglect the systematic learning of domain-specific reasoning patterns and thinking skills.
    \item[P2.] \textbf{Prepared Closed-book Exam}: Conventional pre-training and post-training methods like continual pre-training (CPT) and supervised fine-tuning (SFT) under standard settings \citep{ke2023continual,sun2020ernie}, these approaches incur high training costs for memorization, while knowledge is untraceable and prone to hallucinations.
\end{itemize}

\paragraph{Motivation} 
As Confucius stated, “Learning without thought is labor lost; thought without learning is perilous.” This world-renowned axiom reveals the synergistic relationship between knowledge acquisition and higher-order cognitive processes. Notably, the fields of education and deep learning exhibit profound parallels: domain adaption mirrors subject mastery, training strategies align with pedagogical methodologies, curated datasets correspond to curricular materials, and loss functions reflect educational objectives.
Through the lens of Bloom’s Taxonomy \citep{bloom1964taxonomy,krathwohl2002revision}, the fundamental model of educational objectives, the development of problem-solving abilities requires the harmonious integration of knowledge and cognitive processes. The cognitive processes form a hierarchical structure: from basic knowledge recall to advanced skills like analysis, evaluation, and creation (Fig. \ref{fig:Overview}). The critical insight is: the memorization of massive domain knowledge happens before, competes with and even hinders the learning of higher-order thinking skills, particular within constrained parameter budgets. Thus, the natural research question is: \textit{``Is it possible to decouple and bypass the memorization of domain knowledge, thereby prioritizing and accelerating reasoning modeling?''}

To bridge this crucial gap, this paper introduces Retrieval-Augmented Reasoning Modeling (\textbf{RARE}), as the third paradigm for domain-specific intelligence:
\begin{itemize}
    \item[P3.] \textbf{Prepared Open-book Exam}: RARE skips the parameter-intensive process of knowledge memorization, instead prioritizing the cultivation of reasoning capabilities during training. At inference time, it dynamically retrieves necessary knowledge from external knowledge stores, supported by diverse retrieval mechanisms and infrastructures. This paradigm reallocates model capacity from rote memorization to reasoning-focused parameters, achieving unprecedented cost-effectiveness while maintaining knowledge accuracy and updatability.
\end{itemize}

RARE’s core philosophy centers on the dual principles of \textit{externalizing domain knowledge} and \textit{internalizing domain thinking}. While LLMs exhibit the potential as strong reasoning engines, their limitations in factual precision and interpretability render them unsuitable as standalone knowledge stores. By decoupling knowledge storage (handled via specialized AI databases) from reasoning (optimized through RARE’s training strategy), RARE allocates computational and parametric resources toward optimizing reasoning pathways rather than static factual storage. Specifically, retrieved knowledge is injected into training prompts, reframing knowledge-related losses like memorization errors into application-oriented losses. Beyond simply scaling model parameters, data, or computational resources, RARE opens a new and promising avenue for large reasoning models.

\paragraph{Contributions} The main contributions can be summarized as follows:
\begin{itemize}
    \item \textbf{Problem Formulation}: We conceptualize and formalize the \textit{knowledge-reasoning capacity trade-off} under constrained resources. By drawing parallels with Bloom’s Taxonomy in educational theory and conducting formal mathematical analysis of optimization objectives, we establish theoretical foundations via computational-educational interdisciplinary analysis.
    
    \item \textbf{Method Innovation}: We propose RARE, a novel paradigm that decouples knowledge storage from reasoning modeling. This framework learns reasoning patterns directly while bypassing lower-level knowledge memorization. At inference time, the RARE-trained reasoning engine is integrated with external knowledge store for complete intelligence.
    
    \item \textbf{Experimental Validation}: Extensive experiments confirm RARE’s effectiveness and efficiency across various domains, modalities, benchmarks and backbones. E.g., RARE-trained Llama-3.1-8B achieved 76.8\% and 81.7\% accuracy on PubMedQA and CoVERT, respectively—surpassing retrieval-augmented DeepSeek-R1 (75.4\% and 63.3\%) and GPT-4 (75.2\% and 65.7\%) with trillion parameters. 
\end{itemize}

%% file: Sections/Related_Work.tex
\section{Related Work}
\label{sec:related_work}
\paragraph{Retrieval-Augmented Generation}
Retrieval-augmented generation (RAG) systems enhance LLMs by incorporating external knowledge during inference \citep{zhao2024retrieval}. 
Modern RAG systems advance in knowledge indexing \citep{BM25,2020RAG,wang_qaencoder_2024,bot}, query rewriting \citep{ma2023query, wang2023query2doc}, document compression \citep{xu2023recomp}, retrieval denoising \citep{2410_can_rag_help_reason,dparag}, iterative retrieval \citep{asai2023self,FLARE} and so on, achieving increasingly high retrieval accuracy. E.g., Memory\textsuperscript{3} \citep{yang2024text} introduces explicit memory as a third form of memory. Compared with parameters and text-based retrieval, it externalizes knowledge into a cheaper, retrievable format. However, RAG methods focus on knowledge supplementation rather than reasoning capacity acquisition—retrieved information serves as input augmentation rather than scaffolding for cognitive process development. RARE redefines RAG's role in the training stage: by injecting retrieved knowledge into training prompts, it transforms retrieval contexts into reasoning skill incubators, enabling models to directly learn cognitive patterns from knowledge-anchored examples. This shifts RAG from a post-hoc patch to an integral component of reasoning capability formation.

Notably, our most relevant work is Retrieval-Augmented Fine-Tuning (RAFT) \citep{zhang2024raftadaptinglanguagemodel}. For the imperfect retrieval issue \citep{wang2024astute,liu2025hoprag}, RAFT solely focuses on identifying helpful information from retrieved documents. It learns to mimic the structured output format of teacher models that extract and directly quote sentences, rather than fostering domain thinking—unleashing reasoning capabilities involving higher-order cognitive processes. Moreover, RAFT introduces two hyperparameters related to the proportion of golden and distract documents during training, which is reported to cause training unstability issues \citep{zhang2024raftadaptinglanguagemodel,fleischer2024rag}.
Finally, RAFT relies on teacher model supervision rather than reward signals, and thus cannot be directly applied in reinforcement learning.

\paragraph{Domain LLMs}
Domain-specific LLMs have emerged as critical tools for addressing specialized tasks.
Previous work focuses primarily on knowledge internalization through specialized pretraining \citep{pal2023med} or fine-tuning \citep{pham2024towards} to incorporate relevant expertise. Models like Med-PaLM \citep{pham2024towards}, Med-PaLM 2 \citep{pal2023med}, ClinicalBERT \citep{huang2019clinicalbert} and BioGPT \citep{luo2022biogpt} leverage domain corpora to embed medical knowledge into parameters, while financial LLMs such as BloombergGPT \citep{wu2023bloomberggpt} and FinBERT \citep{araci2019finbert} adopt similar strategies for economic nuances. However, these approaches face inherent limitations: domain knowledge becomes entangled with model parameters, leading to challenges in updating facts \citep{morley2020ethics} and persistent hallucination risks \citep{zhao2024retrieval}. Recent efforts like retrieval-augmented domain models \citep{xu2023retrieval} partially address these issues but retain limited reasoning capabilities. In contrast, RARE fundamentally externalizes domain knowledge, enabling models to focus on reasoning optimization with minimal memorization overhead.

\paragraph{Reasoning LLMs}
Large reasoning models (LRMs) such as OpenAI-o1~\citep{openai2024openaio1card}, DeepSeek-R1~\citep{deepseek-r1}, and Qwen-QwQ~\citep{qwq-32b-preview}, exhibit a paradigm shift toward test-time scaling through long reasoning steps, which enables smaller models to tackle complex tasks by decomposing problems into cognitive chains \citep{wei2022chain,feng2024towards}. For reasoning modeling, previous methods introduce diverse strategies: Monte Carlo Tree Search enhances decision-making through simulation \citep{jiang2024technical}, deliberate error injection improves error correction \citep{ye2024physics}, knowledge distillation of reasoning paths \citep{2412_min_imitate}, etc. For knowledge retrieval, the reasoning capability is often utilized to empower agentic RAG systems \citep{guan2025deeprag,wang2025chain,tran2024rare,li2025search}. For example, Self-RAG \citep{asai2023self} integrates multi-turn search workflows into its reasoning process. It learns when and how to retrieve through curated SFT datasets. Following this, our concurrent works, such as Search-R1 and R1-Searcher~\citep{jin2025search,chen2025research,song2025r1,gao2025synergizing,sun2025zerosearch}, advance with RL training to incentivize common patterns like query planning and reflection for adaptive retrieval. In stark contrast, RARE optimizes the modeling of domain thinking, such as supplement-demand logic chains, via retrieval integration.

%% file: Sections/Method.tex
\section{RARE: Retrieval-Augmented Reasoning Modeling}
\label{sec:method}
In this section, we introduce the RARE framework, elaborating on how RARE bypasses knowledge memorization and cultivates higher-order think skills. We consider tasks that necessitates both domain-specific knowledge and reasoning capabilities for complete evaluation of intelligence.

\paragraph{Problem Formulation} Given the input instruction $x$ such as user query, the objective is to generate chain-of-thoughts response $y$, consisting of interleaved domain knowledge $k$ and domain thinking (or reasoning steps) $r$. For retrieval mechanism, we consider an off-the-shelf retrieval engine $R(\cdot)$ (e.g., BM25 \citep{robertson2009probabilistic}, DPR \citep{karpukhin2020dense}), which returns retrieved knowledge $R(x)$ for each user query $x$. Given any training sample $(x,R(x),y)$, the learning objective, $\min_{\theta} \mathbb{E}\left[\mathcal{L}_{\mathrm{RARE}} \right] $, is to prioritize and accelerate the learning of domain thinking $r$, rather than domain knowledge $k$.

\subsection{Proof of Concept}

As a starting point, we first consider the simplest case, which can be extended to the general scenarios without loss of generality.
Specifically, the chain-of-thoughts response $y$ is modeled as, $y=k \oplus r$, the concatenation of knowledge and reasoning; its generation is also divided into three discrete processes: 
\begin{enumerate}
    \item \textbf{Knowledge Retrieval}: Given the input $x$, an external retrieval system extracts relevant knowledge $R(x)$ from a large-scale knowledge base.
    \item \textbf{Knowledge Integration}: Given $x$ and $R(x)$, the LLM synthesizes domain knowledge $k$ by integrating their intrinsic parametric knowledge with external inputs, following the conditional distribution $k \sim p(k|x, R(x))$. 
    This integration mainly involves \textit{knowledge extraction, or understanding and application, rather than remembering}.
    \item \textbf{Contextualized Reasoning}: 
    The LLM generates reasoning steps $r$ conditioned on $x$, $R(x)$, and the integrated knowledge $k$, adhering to the reasoning distribution $r \sim p(r|x, R(x), k)$.
\end{enumerate}

\paragraph{Learning Objectives} Here we formally analyze and discuss the learning objectives of both vanilla models and RARE-trained models. 

\begin{itemize}
    \item 
    For the vanilla setting, the joint generation distribution is:
    $$
    p_\text{Vanilla}(y|x) = p_\text{Vanilla}(k\oplus r|x) = \underbrace{p_\theta(k|x)}_{\text{Knowledge Remembering}} \cdot \underbrace{p_\theta(r|x,k)}_{\text{Contextualized  Reasoning}}
    $$
    
    The vanilla loss function optimizes both knowledge remembering and contextualized reasoning:
    \begin{align}
    \mathcal{L}_\text{Vanilla} &= -\mathbb{E}_{(x,k,r)}\left[\log p_\theta(k|x) p_\theta(r|x,k)\right]
    \nonumber \\ 
    &= \underbrace{-\mathbb{E}_{(x,k,r)}\left[\log p_\theta(k|x) \right]}_{\text{Loss of Remembering}} +  \underbrace{-\mathbb{E}_{(x,k,r)}\left[\log  p_\theta(r|x,k)\right]}_{\text{Loss of Reasoning}}.\label{eq:vanilla_loss}
    \end{align}
    
    \item 
    For the RARE paradigm, the joint generation distribution is:
    $$
    p_\text{RARE}(y|x,R(x)) = p_\text{RARE}(k\oplus r|x,R(x)) =  \underbrace{p_{\theta}(k|x,R(x))}_{\text{Knowledge Integration}} \cdot \underbrace{p_\theta(r|x,R(x),k)}_{\text{Contextualized  Reasoning}}
    $$
    
    The loss function of RARE optimizes both knowledge integration and contextualized reasoning:
    \begin{align}
    \mathcal{L}_\text{RARE} &= -\mathbb{E}_{(x,k,r)}\left[\log  p_\theta(k|x,R(x))p_\theta(r|x,R(x),k)\right] \nonumber \\ 
    &= \underbrace{-\mathbb{E}_{(x,k,r)}\left[\log  p_\theta(k|x,R(x))\right] }_{\text{Loss of Integration}}
    + \underbrace{-\mathbb{E}_{(x,k,r)}\left[\log  p_\theta(r|x,R(x),k)\right]}_{\text{Loss of Reasoning}}.\label{eq:rare_loss}
    \end{align}
\end{itemize}

\begin{figure}[htbp]
    \centering
    \subfigure[Knowledge Loss Decreases with Retrieval.]{
        \label{Fig.sub.1}
        \includegraphics[width=0.4\textwidth]{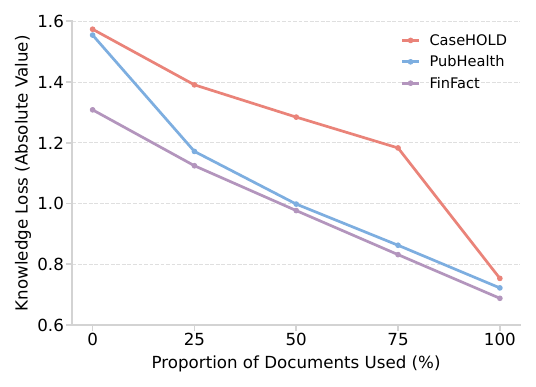}
    }
    \hspace{1em}
    \subfigure[Reasoning Loss Dominates with Retrieval.]{
        \label{Fig.sub.2}
        \includegraphics[width=0.4\textwidth]{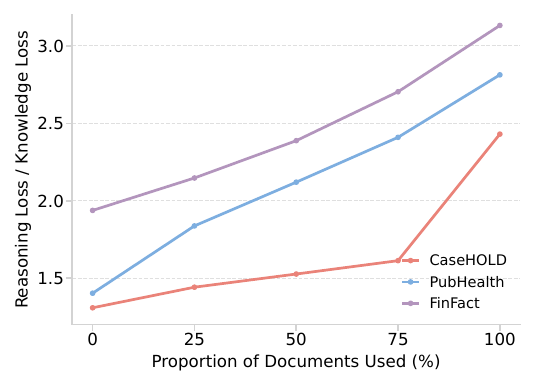}
    }
    \caption{Preliminary experiments of RARE on PubHealth (medical), CaseHOLD (legal), and FinFact (financial). It reveals the dynamics: injecting retrieved context into training prompts transforms learning objectives from  memorization into knowledge integration and contextualized reasoning.}
    \label{fig:preliminary experiment}
\end{figure}

Through the lens of multi-task learning, we compare equation \eqref{eq:vanilla_loss} and \eqref{eq:rare_loss} from two perspectives:

\begin{itemize}
    \item \textbf{Reasoning Augmentation:} Assuming the retrieval quality is high, we have $ p_\theta(k|x,R(x)) \gg  p_\theta(k|x) $, and thus $ p_\theta(k|x,R(x)) \gg  p_\theta(r|x,R(x),k) $ usually. Compared with vanilla models, we express equation \eqref{eq:rare_loss} as:
    \begin{align}
    \mathcal{L}_\text{RARE} 
    = -\mathbb{E}_{(x,k,r)}\left[\log  p_\theta(k|x,R(x))\right]\downarrow  -\mathbb{E}_{(x,k,r)}\left[\log  p_\theta(r|x,R(x),k)\right]\uparrow.
    \label{eq:rare_appro}
    \end{align}
    The arrows indicate the loss function shifts from knowledge to reasoning modeling, which will be further illustrated and quantified in Fig. \ref{fig:preliminary experiment} for the general cases.

    \item \textbf{Knowledge Augmentation:} Comparing the first term of equation \eqref{eq:rare_loss} with equation \eqref{eq:vanilla_loss}, this loss term shifts from knowledge remembering into knowledge integration, such as extracting helpful information from lots of retrieved documents. According to Bloom's taxonomy, \( p_\theta(k|x,R(x)) \) has already stepped into the levels of "understanding" and "application" of knowledge, whereas \( p_\theta(k|x) \) remains at the level of "remembering", i.e. rote memorization. Hence, RARE helps the modeling of knowledge integration, rather than memorization. 
\end{itemize}
\subsection{General Case Extension}
In real-world settings, domain knowledge $k$ and contextualized reasoning $r$ intertwine in an alternative manner (e.g., Fig. \ref{fig:Overview} or step-by-step differential diagnosis in Fig. \ref{fig:Comparsion between RAG and RARE}). The generation process can be extended and modeled as 
$ y_t = \bigoplus_{i=1}^t (k_i \oplus r_i), $ where $k_i$ and $r_i$ represent the knowledge tokens and reasoning tokens at step $ i $, respectively, and $y_t$ represents the concatenation of the autoregressive outputs from the first $ t $ steps. The joint generation distributions remain highly similar:

$$ p_\text{Vanilla}(y_t|x) = \prod_{i=1}^t \underbrace{p_\theta(k_i|x,y_{i-1})}_{\text{Knowledge Remembering}} \cdot \underbrace{p_\theta(r_i|x, y_{i-1},k_i)}_{\text{Contextualized Reasoning}} $$

$$ p_\text{RARE}(y_t|x, R(x)) = \prod_{i=1}^t \underbrace{p_\theta(k_i|x,R(x),y_{i-1})}_{\text{Knowledge Integration}} \cdot \underbrace{p_\theta(r_i|x, R(x), y_{i-1},k_i)}_{\text{Contextualized Reasoning}} $$

\paragraph{Learning Objectives} Here we visualize the learning objectives in the generic and realistic settings. Specifically, we analyze the loss functions on PubHealth, CaseHOLD and FinFact datasets to illustrate the dynamics and effectiveness of RARE. Please refer to Appendix \ref{appendix:implementation} for implementation details. 

\paragraph{Remark} 
For model distillation, cognitive science research suggests that individuals with high cognitive abilities exhibit flexible contextualized adaptation for domain specificity \citep{sa1999domain}. We thus simply adopt the most advanced generic LRMs for domain thinking distillation, and leave more cognitive or educational background in the Appendix \ref{appendix:background}. Notably, the dynamics of RARE discussed within the SFT scenario also apply to RL training seamlessly.
Finally, inference-stage retrieval mechanisms can vary widely, including adaptive or parametric retrieval \citep{asai2023self,yang2024text,su2025parametric}. We adopt the most common single-turn plaintext retrieval for academic simplicity. Further alignment is necessary for other retrieval mechanisms that differ from the training setting, i.e. single-turn plaintext retrieval.

%% file: Sections/Experiments.tex
\section{Experiments}
\label{sec:experiments}
\paragraph{Datasets and Metrics}
For holistic assessments, we mainly assess our models on a diverse set of benchmarks: MedQA \citep{jin2021medqa}, PubMedQA \citep{jin2019pubmedqa}, PubHealth \citep{kotonya2020explainable}, CoVERT \citep{mohr2022covert}, BioASQ \citep{tsatsaronis2015bioasq} (medical), CaseHOLD \citep{zheng2021casehold} (legal), and FinFact \citep{rangapur2023fin} (financial), including medical diagnosis, literature analysis, fact verification and so on. 
Additionally, we incorporate Humanity's Last Exam \citep{phan2025humanity}, VQA-RAD \citep{lau2018dataset}, and MM-RAIT \citep{liu2025benchmarking} into our analysis for an extended examination.
To accurately reflect the actual performance rather than the capacity of format alignment, the answer accuracy is reported.

\paragraph{Implementation and Baselines}
For model distillation, we mainly utilize QwQ-32B as the teacher model, employing rejection sampling with a maximum of 8 attempts to obtain high-quality training data \citep{deepseek-r1}. For medical benchmarks like MedQA, PubMedQA and BioASQ, we retrieve documents from MedOmniKB \citep{chen2025towards}, a multi-source and million-scale medical database, to mimic real-world scenarios with imperfect retrieval. To verify the parameter efficiency of RARE, we consider lightweight LLMs with diverse backbones and enhancements, as well as common practices: CoT and SFT (closed-book settings), RAG, RAFT and SFT+RAG (open-book settings), RARE (proposed method), QwQ-32B, R1-Distill-Llama-8B, GPT-4 and DeepSeek-R1 with and without RAG (common practices). The backbones include Llama-3.1-8B-Instruct, Qwen-2.5-7B-Instruct and Mistral-7B-Instruct-v0.3. We may omit "Instruct" for brevity. Please refer to the Appendix \ref{appendix:implementation} for more details.

\begin{table}[tb]
\caption{Comparing RARE with previous methods across various benchmarks and diverse backbones. We denote the best score in \highlightblue{blue}, and the second-best score in \highlightgreen{green}. Our RARE significantly outperforms other methods, especially on knowledge- and reasoning-intensive tasks.}
\label{table:performance of RARE}
\centering
\resizebox{\textwidth}{!}{
    \begin{tabular}{c|ccccccc}
    \toprule
    \textbf{Method} & \textbf{MedQA} & \textbf{PubMedQA} & \textbf{PubHealth} & \textbf{CoVERT} & \textbf{BioASQ} & \textbf{CaseHOLD} & \textbf{FinFact} \\ 
    \midrule
    \multicolumn{8}{c}{\textbf{Llama-3.1-8B-Instruct}}\\ 
    \midrule 
    CoT & 61.4 & 52.0 & 33.7 & 51.7 & 77.1 & 22.9 & 34.3 \\ 
    SFT & 65.1 & 54.4 & 56.1 & 58.3 & 81.2 & 24.0 & 56.5 \\ 
    RAG & 73.5 & \highlightgreen{72.4} & 50.7 & 57.7 & 86.7 & 62.1 & 53.0 \\ 
    RAFT & 75.9 & 71.2 & 53.2 & \highlightgreen{66.7} & \highlightgreen{91.6} & 33.8 & \highlightgreen{60.5} \\ 
    SFT+RAG & \highlightgreen{83.0} & 69.6 & \highlightgreen{61.4} & 41.7 & 91.6 & \highlightgreen{72.6} & 56.7 \\ 
    \textbf{RARE} & \highlightblue{84.1} & \highlightblue{75.8} & \highlightblue{66.4} & \highlightblue{66.7} & \highlightblue{93.7} & \highlightblue{72.8} & \highlightblue{64.1} \\
    \midrule
    \multicolumn{8}{c}{\textbf{Qwen-2.5-7B-Instruct}}\\ 
    \midrule 
    CoT & 57.5 & 37.2 & 20.0 & 31.7 & 74.3 & 24.5 & 41.7 \\ 
    SFT & 62.2 & 45.8 & 57.2 & 43.3 & 78.8 & 30.2 & 55.8 \\ 
    RAG & 72.7 & 67.8 & 47.0 & 47.0 & \highlightgreen{92.5} & 65.0 & 54.0 \\ 
    RAFT & 75.5 & \highlightgreen{70.3} & 52.0 & \highlightgreen{63.3} & 91.7 & 30.4 & \highlightgreen{61.3} \\ 
    SFT+RAG & \highlightgreen{81.4} & 69.4 & \highlightgreen{63.1} & 53.3 & 91.6 & \highlightgreen{70.4} & 60.2 \\ 
    \textbf{RARE} & \highlightblue{83.0} & \highlightblue{78.6} & \highlightblue{65.1} & \highlightblue{74.1} & \highlightblue{94.0} & \highlightblue{71.1} & \highlightblue{61.6} \\
    \midrule 
    \multicolumn{8}{c}{\textbf{Mistral-7B-Instruct-v0.3}}\\ 
    \midrule 
    CoT & 51.8 & 34.3 & 31.7 & 50.0 & 68.9 & 17.9 & 47.6 \\ 
    SFT & 57.6 & 44.4 & 58.2 & 51.7 & 78.5 & 19.2 & 53.3 \\ 
    RAG & 62.8 & \highlightgreen{64.6} & 48.0 & 46.7 & 87.6 & 41.5 & 58.5 \\ 
    RAFT & 66.7 & 61.0 & 53.1 & \highlightgreen{63.3} & \highlightgreen{90.2} & 34.8 & \highlightgreen{59.1} \\ 
    SFT+RAG & \highlightgreen{75.9} & 59.6 & \highlightgreen{65.0} & 60.0 & 82.7 & \highlightgreen{62.9} & 60.4 \\ 
    \textbf{RARE} & \highlightblue{78.3} & \highlightblue{77.0} & \highlightblue{67.4} & \highlightblue{70.0} & \highlightblue{91.8} & \highlightblue{66.2} & \highlightblue{64.4} \\
    \midrule
    \multicolumn{8}{c}{\textbf{Common Practices}}\\ 
    \midrule
    QwQ-32B & 85.7 & 53.2 & 48.7 & 63.3 & 83.3 & 20.2 & 47.8 \\ 
    QwQ-32B+RAG & 79.1 & 72.4 & 61.2 & 68.3 & 93.2 & 71.9 & 58.9 \\ 
    \midrule
    R1-Distill-Llama-8B & 56.5 & 51.3 & 27.1 & 48.3 & 72.6 & 14.0 & 44.1 \\ 
    R1-Distill-Llama-8B+RAG & 72.7 & 73.5 & 50.1 & 50.0 & 92.6 & 42.0 & 53.1 \\
    \midrule
    GPT-4 & 81.0 & 46.4 & 34.2 & 41.3 & 83.4 & 16.2 & 33.2 \\ 
    GPT-4+RAG & 83.0 & 75.2 & 64.4 & 65.7 & 93.9 & 63.2 & 53.1 \\ 
    \midrule
    DeepSeek-R1 & 93.2 & 53.2 & 49.7 & 65.0 & 87.7 & 20.6 & 49.2 \\ 
    DeepSeek-R1+RAG & 92.8 & 75.4 & 61.3 & 63.3 & 93.9 & 74.1 & 61.3 \\ 
    \bottomrule 
    \end{tabular}
}
\end{table}

\subsection{RARE Achieves Better Accuracy, Efficiency and Robustness}

\paragraph{Reasoning Accuracy}
Extensive experiments demonstrate that RARE achieves state-of-the-art performance across various benchmarks. As shown in Table \ref{table:performance of RARE}, RARE significantly surpasses other models within the same scale. For instance, on the PubMedQA benchmark, RARE achieves accuracy rates of 75.8\%, 76.7\%, and 77.0\% using Llama-3.1-8B-Instruct, Qwen-2.5-7B-Instruct, and Mistral-7B-Instruct-v0.3 respectively, outperforming SFT+RAG (69.6\%, 69.4\%, 59.6\%), RAFT  (71.2\%, 70.3\%, 61.0\%), RAG (72.4\%, 67.8\%, 64.6\%) and other baselines significantly.
\begin{wraptable}{r}{0.5\textwidth}
\centering
\vspace{-1em}
\caption{
Accuracies on multi-modal benchmarks, VQA-RAD (medical) and MM-RAIT (open-domain). RARE (Multi-Task) is fine-tuned on both VQA-RAD and text-only medical datasets. 
}
\label{table:multi-modal}
\resizebox{0.5\textwidth}{!}{
\begin{tabular}{l|cc}
\toprule
    \textbf{Method} & \textbf{VQA-RAD} & \textbf{MM-RAIT} \\ \midrule
    Qwen2.5-VL-7B-Instruct & ~     & ~     \\
    +RAG              & 52.2  & 61.9  \\ 
    +RARE             & 57.8  & \highlightblue{70.9}  \\ 
    +RARE (Multi-Task)& \highlightgreen{61.4}  & -     \\ \midrule
    Qwen2.5-VL-32B-Instruct+RAG & 54.6 & 62.0 \\  
    GPT-4o-mini+RAG & 51.0 & 60.3 \\
    GPT-4o+RAG         & \highlightblue{68.5}    & \highlightgreen{67.5}     \\
    \bottomrule
\end{tabular}
}
\vspace{-1em}
\end{wraptable}
Table~\ref{table:multi-modal} confirms RARE's effectiveness on multi-modal scenarios with Qwen2.5-VL-32B-Instruct as the teacher model. For example, on open-domain benchmark MM-RAIT, RARE improves over RAG by 9.0\% (from 61.9\% to 70.9\%) while GPT-4o+RAG attains only 67.5\%; on medical benchmark VQA-RAD, RARE (Multi-Task), fine-tuned on both VQA-RAD and text-only medical datasets, achieves 61.4\% accuracy and surpasses RAG by 9.2\%. Therefore, RARE not only enhances the synergy between domain knowledge and domain thinking, but also prioritize the learning of domain thinking, which is more transferable across different tasks.

\paragraph{Reasoning Efficiency} RARE introduces a groundbreaking improvement in cost-effectiveness. By externalizing knowledge retrieval, RARE reallocates model capacity from parameter-heavy memorization to reasoning-focused parameters, allowing lightweight architectures to achieve state-of-the-art performance with minimal deployment cost and inference latency \citep{bai2024beyond,samsi2023words}. For example, RARE-trained Llama-3.1-8B achieves 84.1\% and 93.7\% accuracies on MedQA and BioASQ respectively, while GPT-4+RAG attains only 83.0\% and 93.9\% with substantial computational resources. For reasoning models, RARE not only surpasses the teacher model QwQ-32B+RAG, but also beats the most advanced DeepSeek-R1+RAG. Specifically, RARE-trained Llama-3.1-8B achieves 66.4\%, 72.8\% and 64.1\% on PubHealth, CaseHOLD and FinFact respectively, while DeepSeek-R1+RAG attains 61.3\%, 74.1\% and 61.3\% with trillion parameters. These astonishing results highlight RARE's ability to decouple knowledge storage from reasoning optimization, enabling lightweight models to rival or exceed large-scale systems. Please refer to Fig. \ref{fig:Benchmark} for more intuitive landscape.

\begin{wrapfigure}{r}{0.5\textwidth}
\vspace{-1em}
\centering
\includegraphics[width=0.4\textwidth]{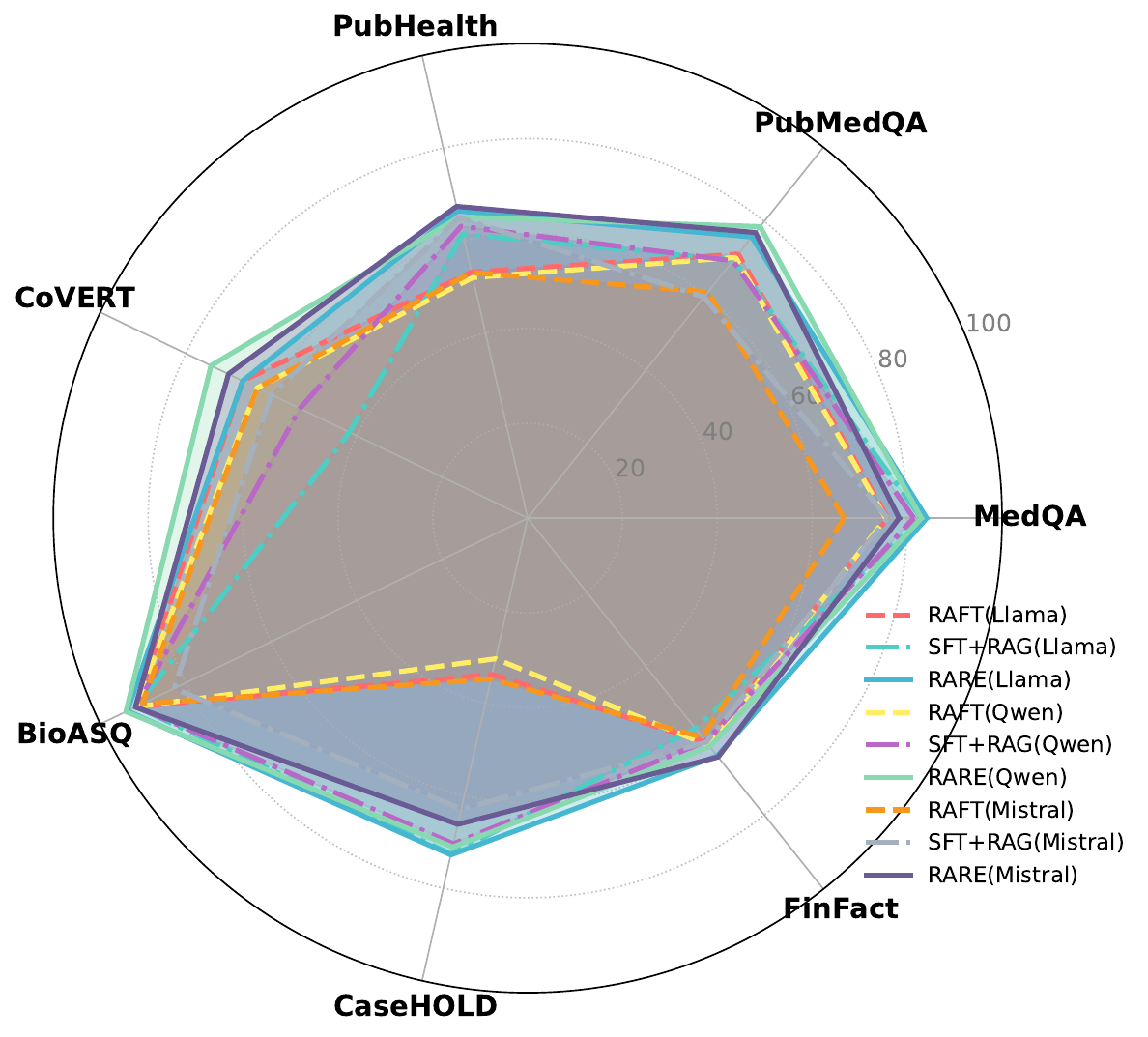}
\caption{
Comparing RARE with RAFT and SFT+RAG across various benchmarks and diverse backbones. RARE not only achieves better accuracy, but also exhibits better training robustness.}
\label{figure:radar}
\vspace{-1em}
\end{wrapfigure}
\paragraph{Reasoning Robustness} Additionally, RARE exhibits enhanced robustness compared with other training strategies. First, as shown in Fig. \ref{figure:radar}, RARE's performance remains stable across various backbones and benchmarks, while RAFT and SFT+RAG suffer from training unstability issues \citep{fleischer2024rag, zhang2024raftadaptinglanguagemodel}. For example, on the PubMedQA benchmark, SFT+RAG drops behind RAG by 3\% (from 72.4\% to 69.6\%) for Llama-3.1-8B, whereas RARE consistently outperforms across backbones without significant fluctuations; on the CoVERT benchmark, SFT+RAG achieves 60.0\% accuracy for Mistral-7B-v0.3 but suffers an 18\% drop when applied to Llama-3.1-8B. In contrast, RARE maintains 66.7\% and 70.0\% accuracies respectively. The robustness advantage of RARE is particularly evident on the CaseHOLD benchmark, where RAFT's performance is notably poor, as clearly shown in Fig. \ref{figure:radar}. Moreover, we observed that the performance of SFT+RAG-trained Llama-3.1-8B on MedQA shows unstable training performance, ranging from 74\% to 83\%. For more fair comparison, we still reported the peak performance for baselines. Finally, RARE is also robust to retrieval quality. Even in imperfect retrieval settings like PubMedQA, RARE can still identify and utilize helpful knowledge from incomplete or noisy information, thanks to the learning objective of knowledge integration and contextualized reasoning.

\subsection{Analysis and Discussion}

\paragraph{Multi-Task and Cross-Task Learning}
As shown in Tables \ref{table:multi_task} and \ref{table:hle}, RARE-trained models achieve comparable or superior accuracy in multi-task settings compared to single-task models. For instance, Llama3.1-8B+RARE improves CoVERT accuracy from 66.7\% (single-task) to 75.0\% (cross-task), suggesting that reasoning patterns learned through RARE transfer effectively across related tasks. The cross-task performance on the challenging Humanity’s Last Exam (HLE) benchmark further underscores this point. While all models struggle with HLE, RARE-trained Llama3.1-8B achieves 12.2\% accuracy, on par with GPT-4+RAG (12.24\%) and DeepSeek-R1+RAG (14.29\%).

\begin{table}[H]
    \centering
    \caption{
     Comparative accuracy of RARE using single-task, multi-task, and cross-task learning. Results are presented in the format ``single-task/multi-task'' or ``cross-task''. RARE demonstrates robust performance in multi-task settings, even surpassing single-task models. 
     Refer to Table \ref{table:hle} for baseline results on HLE (Bio/Med).
    }
    \label{table:multi_task}
    \resizebox{\textwidth}{!}{
    \begin{tabular}{c|cccccc}
    \toprule
        \textbf{Method} & \textbf{MedQA} & \textbf{PubMedQA} & \textbf{PubHealth} & \textbf{CoVERT} & \textbf{BioASQ} & \textbf{HLE (Bio/Med)} \\ \midrule
        Llama3.1-8B-Instruct+RARE & 84.1/83.4 & 75.8/75.8 & 66.4/66.5 & 66.7/75.0 & 93.7/93.0 & 12.2 \\ 
        Qwen-2.5-7B-Instruct+RARE & 83.0/82.3 & 78.6/74.2 & 65.1/64.0 & 74.1/76.7 & 92.9/92.7 & 11.6 \\
        Mistral-7B-Instruct-v0.3+RARE & 78.3/80.6 & 77.0/75.6 & 67.4/66.8 & 70.0/66.7 & 91.8/91.9 & 12.2 \\ \bottomrule
    \end{tabular}
    }
\end{table}
\begin{table}[H]
    \centering
    \caption{
    Accuracies of state-of-the-art models on the Humanity’s Last Exam for Biology/Medicine domain. Results highlight the challenges of achieving high performance on this hard benchmark.
    }
    \label{table:hle}
    \resizebox{\textwidth}{!}{
    \begin{tabular}{c|ccccccccccc}
    \toprule
        \textbf{Model} & \textbf{GPT-4o} & \textbf{GPT-4o+RAG} & \textbf{DeepSeek-R1} & \textbf{DeepSeek-R1+RAG} & \textbf{o3-mini (high)} & \textbf{QwQ-32B} & \textbf{QwQ-32B+RAG} & \textbf{R1-Distill-Llama-8B+RAG} \\ \midrule
        ACC & 7.48 & 12.24 & 13.61 & 14.29 & 8.16 & 7.48 & 12.24 & 4.76 \\
    \bottomrule
    \end{tabular}
    }
\end{table}

\paragraph{Parameter-Efficient Fine-Tuning}
Table \ref{table:peft} illustrates that, RARE maintains strong performance under parameter-efficient fine-tuning (PEFT) with LoRA. With only 1.03\% learnable parameters (rank=64), RARE achieved an accuracy of 82.3\% on MedQA and 76.6\% on PubMedQA, nearly matching full RARE's performance of 84.1\% and 75.8\% respectively. This efficiency stems from its core mechanism: by offloading knowledge storage, RARE prioritizes reasoning learning under limited parameter budgets.
Notably, modern LoRA deployment techniques \citep{sheng2023slora, yu2025ssmlora} enable hosting thousands of LoRA adapters concurrently, making it possible to building generalist LLMs equipped with multiple domain-specific intelligences.

\begin{table}[htb]
    \centering
    \caption{ Accuracies of RARE with LoRA parameter-efficient fine-tuning (r=32, r=64, r=128) on Llama-3.1-8B-Instruct. RARE still exhibits satisfactory accuracy with modest amount learnable parameters (e.g. r=64), thanks to prioritizing reasoning learning under limited parameter budgets. }
    \label{table:peft}
    \resizebox{\textwidth}{!}{
    \begin{tabular}{l|ccccccc}
    \toprule
        \textbf{Method} & \textbf{MedQA} & \textbf{PubMedQA} & \textbf{PubHealth} & \textbf{Covert} & \textbf{BioASQ} & \textbf{CaseHOLD} & \textbf{FinFact} \\ \midrule
        RAG & 73.5 & 72.4 & 50.7 & 57.7 & 86.7 & 62.1 & 53.0 \\ 
        RARE+LoRA (r=32, 0.52\%) & 81.0 & 75.4 & 64.7 & 65.0 & 90.7 & 71.2 & 61.6 \\ 
        RARE+LoRA (r=64, 1.03\%) & 82.3 & \textbf{\highlightblue{76.6}} & 65.3 & 65.5 & \highlightgreen{93.5} & \textbf{\highlightblue{73.4}} & \highlightgreen{62.0} \\ 
        RARE+LoRA (r=128, 2.05\%) & \highlightgreen{82.6} & \highlightgreen{76.0} & \highlightgreen{65.5} & \highlightgreen{66.7} & 93.4 & 72.6 & 61.0 \\ 
        RARE (full parameter)  & \textbf{\highlightblue{84.1}} & 75.8 & \textbf{\highlightblue{66.4}} & \highlightblue{66.7} & \textbf{\highlightblue{93.7}} & \highlightgreen{72.8} & \textbf{\highlightblue{64.1}} \\ 
    \bottomrule
    \end{tabular}
    }
\end{table}

\begin{wraptable}{r}{0.5\textwidth}
\begin{minipage}{0.5\textwidth}
\centering
\vspace{-1em}
\caption{ 
Performance of RARE with KTO reinforcement learning on Llama-3.1-8B-Instruct. RARE's principles are effective for RL training, which provides promising further improvements.
}
\label{table:rl_part}
\resizebox{\textwidth}{!}{
\begin{tabular}{l|cc}
\toprule
    \textbf{Method} & \textbf{PubHealth} & \textbf{CoVERT} \\ \midrule
    Llama3.1-8B-Instruct & ~     & ~     \\
    +RAG & 50.7  & 57.7  \\ 
    +RARE (SFT) & 66.4  & \highlightgreen{66.7}  \\ 
    +RARE (KTO)& \highlightblue{69.2}  & 65.0     \\ 
    +RARE (KTO based on SFT) & \highlightgreen{66.9} & \highlightblue{81.7} \\             	              
    \bottomrule
\end{tabular}
}
\vspace{-1em}
\end{minipage}
\end{wraptable}
\paragraph{Reinforcement Learning of RARE} Table \ref{table:rl_part} shows enhanced performance of RARE with reinforcement learning. We adopt Kahneman-Tversky Optimization (KTO) \citep{ethayarajh2024kto} for the simplicity. 
For example, on the PubHealth dataset, Llama-3.1-8B directly trained with KTO achieved an accuracy of 69.2\%, which is higher than the 66.4\% achieved with SFT. Similarly, on the CoVERT dataset, the accuracy increased from 66.7\% with SFT to 81.7\% when using KTO based on SFT. These promising results not only validate RARE’s compatibility with reinforcement learning, but also solidify RARE's position as a powerful paradigm for domain-specific intelligence.

Due to the page limit, we leave the discussion on data-efficient fine-tuning, rejection sampling, and several case studies with aha-moments in the Appendix for interested readers.

%% file: Sections/Conclusion.tex
\section{Conclusion}
\label{sec:conclusion}
In this paper, we introduced RARE, a paradigm-shifting framework that redefines domain-specific intelligence by decoupling knowledge storage from reasoning optimization. By injecting retrieved knowledge into training prompts with masked losses, this simple-yet-effective approach enables lightweight models to bypass parameter-intensive memorization and prioritize higher-order cognitive skills. Extensive experiments across medical, legal, financial, open-domain and multi-modal benchmarks demonstrate that RARE-trained models achieve state-of-the-art performance, surpassing retrieval-augmented GPT-4 and DeepSeek-R1 by up to approximately 20\% in accuracy. Future work should explore applying RARE to more inference-stage retrieval or memory mechanisms, diverse RL algorithms, agentic LLMs with tool use, and generalist LLMs with multiple domain-specific intelligences. The success of RARE underscores the importance of rethinking the balance between memorization and reasoning. We hope RARE will inspire both academic and industrial communities and serve as a cornerstone for next-generation AI systems.

%% file: Sections/Appendix.tex
\clearpage
\appendix
\section*{Appendix}
\label{sec:appendix}
\startcontents[sections]
\printcontents[sections]{l}{1}{\setcounter{tocdepth}{3}}

\newpage
\section{More Experiments and Discussions}

\subsection{Extended Performance Landscape}
\begin{figure}[H]
     \centering
     \includegraphics[width=\textwidth]{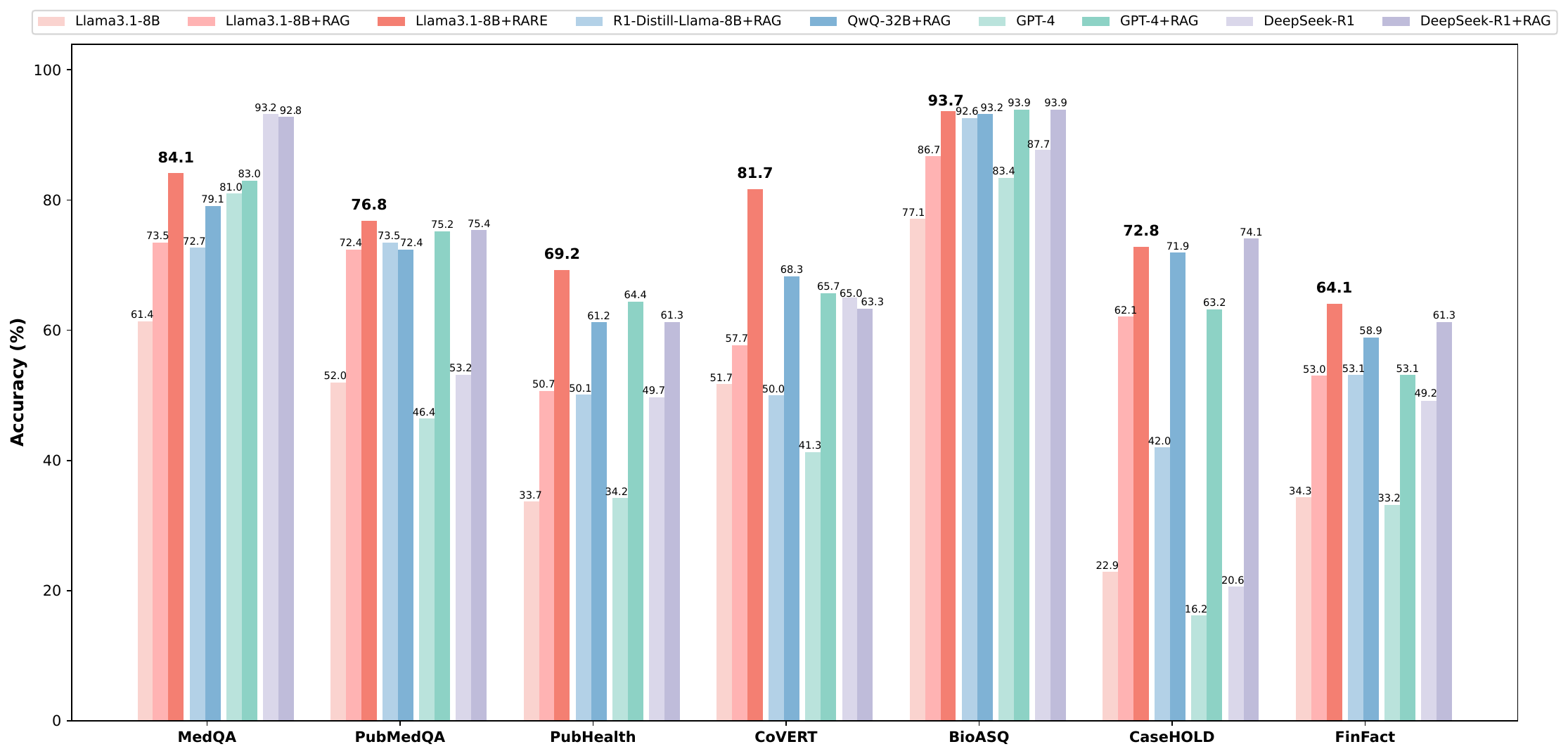}
     \caption{Performance of RARE versus baselines on benchmarks (extended version).}
    \label{fig:benchmark  (extended version)}
\end{figure}

\subsection{Data-Efficient Fine-Tuning}

To better demonstrate the performance of RARE, we evaluated its data efficiency compared to SFT+RAG. To ensure a fair comparison, we selected samples from the intersection of both methods after rejection sampling as the complete training dataset. Based on this foundation, we conducted training with 0\%, 20\%, 50\%, 80\% and 100\% dataset. Given consistent and high-quality training samples, we were able to demonstrate the superior data efficiency of RARE.

\begin{figure}[h]
\centering
\includegraphics[width=0.9\linewidth]{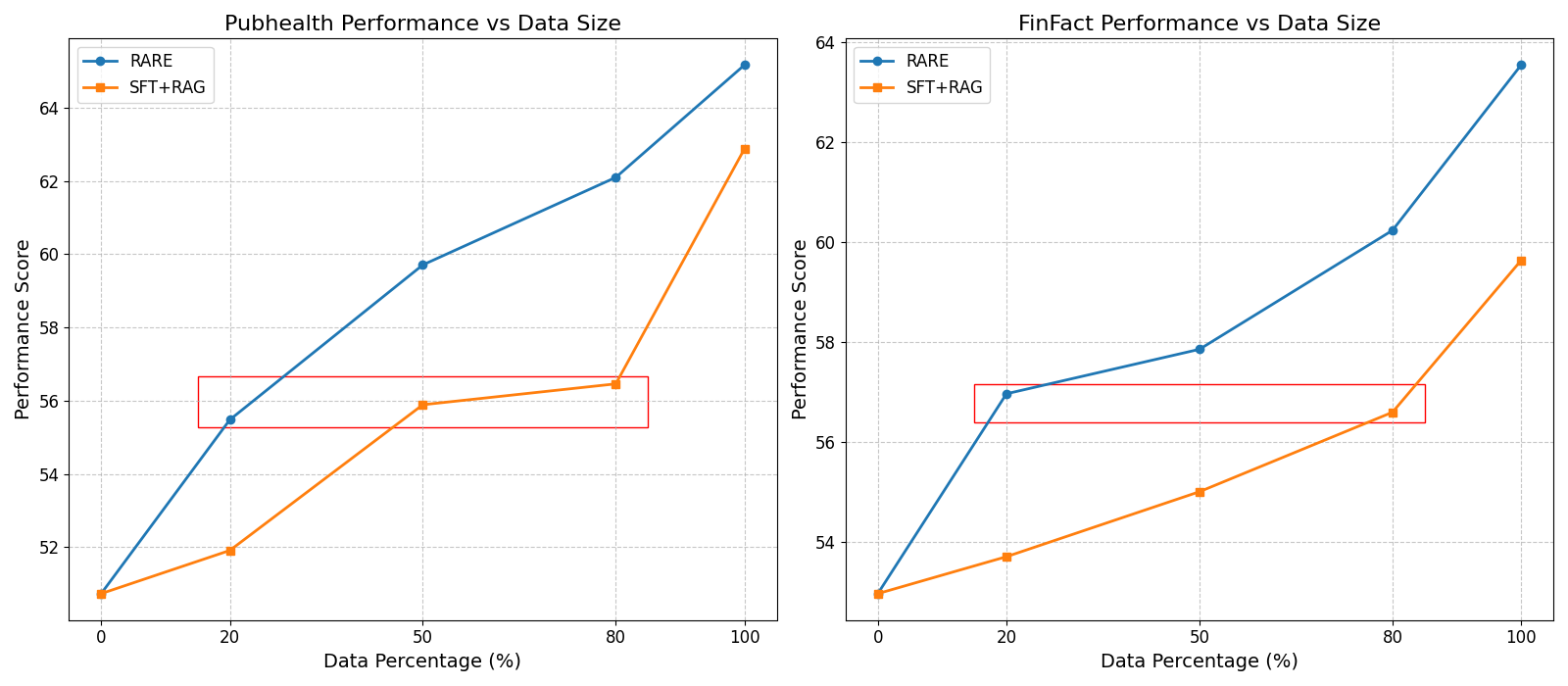}
\caption{As the amount of training data increases, the performance trends of the RARE and SFT+RAG methods on the FinFact and PubHealth datasets. Notably, we have highlighted with red frames that RARE achieves comparable ane even better performance to SFT+RAG when utilizing only 20\% of the training data volume, whereas SFT+RAG requires 80\% of the data volume.}
\label{fig:data_efficiency}
\end{figure}

As shown in Figure~\ref{fig:data_efficiency}, RARE consistently outperforms the SFT+RAG method across all data scales. The figure clearly demonstrates that RARE can achieve with merely 20\% of the training data what SFT+RAG requires 80\% to accomplish. Furthermore, when both methodologies utilize identical data quantities, RARE exhibits significantly superior performance gains. To verify that this advantage is not coincidental, we conducted statistical error testing. The t-test between the two experimental groups yielded a p-value of only 0.0007 on the FinFact dataset and 0.0115 on the PubHealth dataset (both p-value < 0.05, indicating statistical significance). These results demonstrate that the RARE method possesses significant advantages in data efficiency. From a graphical perspective, the performance improvement curve for RARE exhibits pronounced enhancement with increasing data volume without demonstrating convergence, further validating the necessity of distilling the teacher model through repeated rejection sampling to generate additional high-quality data.

\begin{table}[h]
\centering
\caption{Performance comparison between RARE and SFT+RAG across different training data scales}
\label{tab:data_efficiency}
\begin{tabular}{l|ccccc}
\hline
\multirow{2}{*}{Method} & \multicolumn{5}{c}{Data Scale} \\
 & 100\% & 80\% & 50\% & 20\% & 0\% \\
\hline
\multicolumn{6}{c}{FinFact Dataset} \\
\hline
RARE & 63.6 & 60.2 & 57.9 & 57.0 & 53.0 \\
SFT+RAG & 59.6 & 56.6 & 55.0 & 53.7 & 53.0 \\
\hline
\multicolumn{6}{c}{PubHealth Dataset} \\
\hline
RARE & 65.2 & 62.1 & 59.7 & 55.5 & 50.7 \\
SFT+RAG & 62.9 & 56.5 & 55.9 & 51.9 & 50.7 \\
\hline
\end{tabular}
\end{table}

The specific experimental data presented in Table~\ref{tab:data_efficiency} confirms that the RARE methodology maintains a substantial advantage across all data scales in both datasets, while simultaneously validating the significance of our rejection sampling strategy. Overall, this series of experiments reveals the potential of RARE in limited-data scenarios, which has significant implications for practical applications.

\subsection{Impact of Rejection Sampling}
\label{appendix:Rejection Sampling}
We discovered that when distilling teacher model, employing a rejection sampling strategy effectively improves the quality and quantity of training set, thereby enhancing post-training results. Specifically, our strategy allows the teacher model eight response opportunities for each sample. If the model answers correctly within these eight attempts, no further distillation is performed; conversely, if all eight responses are incorrect, we preserve the most concise response. As shown in Table~\ref{tab:sampling_comparison}, after repeated sampling, the accuracy of the QwQ-32B+RAG demonstrated significant improvement across all datasets, substantially enhancing data quality.

\begin{table}[htbp]
\centering
\caption{Accuracy comparison between QwQ-32B+RAG with different sampling counts}
\label{tab:sampling_comparison}
\begin{tabular}{l|ccccccc}
\hline
Method & MedQA & PubMedQA & PubHealth & CoVERT & BioASQ & CaseHold & FinFact \\
\hline
Sampling1 & 78.04 & 72.60 & 72.05 & 67.50 & 94.43 & 72.22 & 57.69 \\
Sampling8 & 88.38 & 82.00 & 77.20 & 78.75 & 95.29 & 83.33 & 67.76 \\
\hline
\end{tabular}
\end{table}

For the distilled training data, we implemented different processing approaches based on dataset size. For datasets with original sizes exceeding 2k samples (MedQA, PubHealth, CaseHold, FinFact), we utilized only correctly answered samples for training. For PubMedQA, CoVERT, and BioASQ, we employed the complete training set. 
We conducted experiments to validate the effectiveness of our rejection sampling strategy. Table~\ref{tab:rare_comparison} compares the results after training for 5 epochs on different base models using training sets  with and without rejection sampling.

\begin{table}[htbp]
\centering
\caption{Performance comparison of different base models with RARE method across datasets}
\label{tab:rare_comparison}
\begin{tabular}{l|ccccc}
\hline
Method & MedQA & PubMedQA & PubHealth & CoVERT & BioASQ \\
\hline
\multicolumn{6}{l}{Llama3.1-8B} \\
\hline
RARE (sampling 1) & 80.6 & 75.0 & 63.4 & 66.7 & 93.2 \\
RARE (sampling 8) & 84.1 & 75.8 & 66.4 & 66.7 & 93.7 \\
\hline
\multicolumn{6}{l}{Qwen2.5-7B} \\
\hline
RARE (sampling 1) & 81.2 & 76.7 & 63.1 & 68.3 & 92.9 \\
RARE (sampling 8) & 83.0 & 78.6 & 65.1 & 74.1 & 94.0 \\
\hline
\multicolumn{6}{l}{Mistral-7B-v0.3} \\
\hline
RARE (sampling 1) & 71.3 & 76.9 & 64.9 & 67.8 & 91.0 \\
RARE (sampling 8) & 78.3 & 77.0 & 67.4 & 70.0 & 91.8 \\
\hline
\end{tabular}
\end{table}

\subsection{Case Study \& Aha-Moment}
Here, we present several case studies for more intuitive comparison. As illustrated in Tables \ref{tab:case-study-part1} and \ref{tab:case-study-part2}, we analyze sample "medqa-000447" from the MedQA dataset, comparing outputs from four methodologies: RAG, SFT+RAG, RAFT, and RARE.

\newcommand{\thinkrow}[1]{\rowcolor{blue!10} #1 \\}
\newcommand{\sourcerow}[1]{\rowcolor{green!10} #1 \\}

\begin{table}[htbp]
  \caption{
  The sample (ID: "medqa-000447") from the MedQA benchmark with standard answer, comparing the outputs of RAG, SFT+RAG, RAFT and RARE (Part I). \colorbox{blue!10}{$\phantom{text}$} represents Contextualized Reasoning, \colorbox{green!10}{$\phantom{text}$} indicates Knowledge Integration, and \colorbox{yellow!10}{$\phantom{text}$} signifies the final answer.
  }
  \label{tab:case-study-part1}
  \begin{center}
  \begin{tabular}{p{0.95\textwidth}}
  \toprule
  \textbf{Question:} A 52-year-old man with heavy alcohol use presents with bloody vomiting, hypotension (85/59 mmHg), tachycardia (122/min), spider angiomata, and distended abdomen. Which is most likely decreased compared to a healthy adult?\\
  A. Systemic vascular resistance\\
  B. Fractional tissue oxygen extraction\\
  C. Carotid baroreceptor firing activity\\
  D. Cardiac inotropy \\
  \textbf{Answer:} C \\
  \midrule 
  \rowcolor{gray!20} \multicolumn{1}{c}{\textbf{RAG}} \\
  \thinkrow{<think> Okay, let's tackle this question step by step... The patient is a 52-year-old man with a history of heavy alcohol use, presenting with vomiting, hypotension, tachycardia, and signs of shock...}
  \sourcerow{From retrieved documents: In septic shock, SVR is typically decreased due to vasodilation caused by cytokines...}
  \thinkrow{Key considerations: 1. Systemic Vascular Resistance (SVR): In septic shock, SVR is typically decreased due to vasodilation... The patient's presentation aligns with sepsis...}
  \thinkrow{2. Fractional Tissue Oxygen Extraction (B): In shock, the body increases oxygen extraction to maintain oxygen delivery, so this is likely increased, not decreased...}
  \thinkrow{3. Carotid Baroreceptor Firing Activity (C): Hypotension activates baroreceptors, increasing their firing activity to stimulate sympathetic responses... (Note: This reasoning is incorrect)}
  \thinkrow{4. Cardiac Inotropy (D): While sepsis depresses myocardial contractility over time, the question focuses on the current condition...}
  \thinkrow{The most consistent finding with septic shock (a common cause in liver disease) is decreased SVR.</think>}
  \rowcolor{yellow!10} <answer>A</answer> \\
  \midrule
  \rowcolor{gray!20} \multicolumn{1}{c}{\textbf{SFT+RAG}} \\
  \thinkrow{<think> Okay, let's tackle this question step by step. The patient is a 52-year-old man with a history of heavy alcohol use, presenting with vomiting, hypotension, tachycardia, and signs of shock...}
  \sourcerow{In distributive shock (like septic or neurogenic shock), SVR is typically low...}
  \thinkrow{The patient presents with distributive shock: hypotension with tachycardia, vasodilation signs... The spider angiomata might indicate chronic liver disease...}
  \thinkrow{Looking at the options: A. Systemic vascular resistance (SVR): In distributive shock, SVR is typically low...}
  \thinkrow{C. Carotid baroreceptor firing activity: In hypotension, they would fire more to stimulate increased heart rate and vasoconstriction... (Note: This reasoning is incorrect)}
  \thinkrow{The question asks which is DECREASED. So A (SVR) is decreased... The answer is A.</think>}
  \rowcolor{yellow!10} <answer>A</answer> \\
  \bottomrule
  \end{tabular}
  \end{center}
  \end{table}

This clinical vignette from the MedQA dataset presents a complex medical reasoning task, describing a patient's clinical manifestations and vital signs, and requiring the model to determine which physiological parameter is most likely decreased. \textit{Notably, Llama3.1-8B-Instruct with RAG, SFT+RAG, and RAFT all incorrectly selected option A, while the RARE-trained model initially considered A but ultimately arrived at the correct answer C through reflection and evidence-based medicine \citep{sackett1997evidence}}.

  \begin{table}[htbp]
  \caption{
    The sample (ID: "medqa-000447") from the MedQA benchmark with standard answer, comparing the outputs of RAG, SFT+RAG, RAFT and RARE (Part II). \colorbox{blue!10}{$\phantom{text}$} represents Contextualized Reasoning, \colorbox{green!10}{$\phantom{text}$} indicates Knowledge Integration, and \colorbox{yellow!10}{$\phantom{text}$} signifies the final answer. Notably,
    the Aha-moment is in \textcolor{red}{\textbf{red}}, which adheres to the \textit{Evidence-Based Medicine} process \citep{sackett1997evidence}.
  }
  \label{tab:case-study-part2}
  \begin{center}
  \begin{tabular}{p{0.95\textwidth}}
  \toprule
  \rowcolor{gray!20} \multicolumn{1}{c}{\textbf{RAFT}} \\
  \thinkrow{{\texttt{\#\#\#Reason:}} The patient in this scenario is exhibiting signs of septic shock, which is characterized by low blood pressure, tachycardia, altered mental status, and decreased peripheral vascular resistance...}
  \thinkrow{In septic shock, there is often a decrease in systemic vascular resistance (SVR). This is due to the effects of vasodilatory mediators such as nitric oxide, which are released in response to systemic inflammation...}
  \sourcerow{"{\texttt{\#\#begin\_quote\#\#}} In septic shock, a MAP of 65 mmHg is considered sufficient for end-organ perfusion. {\texttt{\#\#end\_quote\#\#}}"}
  \sourcerow{"{\texttt{\#\#begin\_quote\#\#}} In hyperdynamic septic shock, tachycardia is present, the cardiac output is normal or elevated, and the systemic vascular resistance is reduced. {\texttt{\#\#end\_quote\#\#}}"}
  \sourcerow{"{\texttt{\#\#begin\_quote\#\#}} NO and peroxynitrite are mainly responsible for vasoplegia and vascular hyporeactivity. {\texttt{\#\#end\_quote\#\#}}"}
  \thinkrow{Given these considerations, the most likely decreased parameter in this patient compared to a healthy adult is systemic vascular resistance.}
  \rowcolor{yellow!10} <answer>A</answer> \\
  \midrule
  \rowcolor{gray!20} \multicolumn{1}{c}{\textbf{RARE}} \\
  \thinkrow{<think> Okay, let's tackle this question step by step...}
  \thinkrow{The patient's symptoms suggest upper GI bleeding given the black, tarry vomitus... The distended abdomen and spider angiomata might point towards liver issues like cirrhosis... The hypotension and tachycardia are signs of shock, likely hypovolemic due to blood loss.}
  \sourcerow{In septic and anaphylactic shock, there's a decrease in SVR due to cytokines... But in hypovolemic shock, the body would try to increase SVR via sympathetic response...}
  \thinkrow{In hypovolemic shock from blood loss, the body's response is to compensate by increasing SVR through vasoconstriction... But in this case, the patient is in shock from blood loss...}
  \thinkrow{In cirrhosis with hypovolemia, SVR increases from baseline but remains lower than a healthy person's... Therefore, A is decreased.}
  \sourcerow{In the "Harrison's" text, in septic shock's hyperdynamic phase, SVR is reduced. But this is septic, not hypovolemic...}
  \thinkrow{Fractional tissue oxygen extraction would be increased, not decreased... Baroreceptors respond to changes in BP. The patient's BP is low (85/59)... In hypovolemic shock, the heart's contractility might be normal or even increased...}
  \thinkrow{Suppose a healthy person has an SVR of 1200 under normal conditions. In hypovolemia, their SVR increases to 1800... A patient with cirrhosis has a lower SVR, say 800, but then in hypovolemia, their SVR increases to 1200...}
  \thinkrow{Therefore, A (SVR decreased) would be correct.}
  \thinkrow{\textcolor{red}{\textbf{Wait, but I'm confused}}... Let me think again. The patient's BP is 85/59. A healthy adult in hypovolemia might have a lower BP but not as low...}
  \sourcerow{From "Physiology, Systemic Vascular Resistance -- Pathophysiology": In septic and anaphylactic shock, SVR is decreased. In hypovolemic shock, the body increases SVR...}
  \thinkrow{So the patient here (hypovolemic from GI bleed) would have increased SVR. Therefore, A is not decreased. Baroreceptor firing (C) is decreased because BP is low...}
  \thinkrow{Therefore, the correct answer is C.</think>}
  \rowcolor{yellow!10} <answer>C</answer> \\
  \bottomrule
  \end{tabular}
  \end{center}
  \end{table}

The base model employing standard RAG demonstrated comprehensive reasoning, yet exhibited a fundamental misconception by misclassifying the patient as experiencing septic shock rather than hypovolemic shock. Despite systematically evaluating all options, it fundamentally misunderstood baroreceptor physiology, erroneously assuming that hypotension would increase rather than decrease baroreceptor firing. This conceptual error led to the premature elimination of the correct option C, redirecting focus toward options A and D.

The SFT+RAG approach enhanced the model's reasoning efficiency, producing more concise deliberation while maintaining similar logical pathways. However, it perpetuated the same conceptual misclassification as RAG, incorrectly categorizing the patient's condition as distributive shock and misinterpreting baroreceptor function. Although fine-tuning improved reasoning efficiency, it failed to develop the capacity of sophisticated knowledge integration for complex medical reasoning.

The RAFT demonstrated a distinctive approach, structurally separating reasoning from knowledge integration. This framework follows a more direct pattern: presenting reasoning, citing supporting literature, and formulating conclusions. However, this separation potentially introduces vulnerability to confirmation bias—exemplified in this case by the model's presupposition of septic shock, followed by selective citation of literature supporting this misdiagnosis. This illustrates the "Texas sharpshooter fallacy" where evidence is selectively marshaled to support a predetermined conclusion.

In stark contrast, the RARE-trained model exhibited superior medical reasoning capabilities characterized by intellectual humility and epistemic vigilance. It deeply integrated external knowledge with contextual reasoning while continuously evaluating its own conclusions. The highlighted "Aha moment" demonstrates this metacognitive capability—when faced with conflicting inferences, the model revisited foundational concepts, reconsidered diagnostic categorization, and ultimately arrived at the correct diagnosis of hypovolemic shock. RARE represents the only methodology that successfully identified and corrected the misconception regarding carotid baroreceptor function in hypotension.

RARE's intellectual humility, manifested through ongoing self-critique and willingness to revise initial hypotheses, represents a fundamental advancement toward more reliable domain decision support systems. This ability to systematically question assumptions and reconcile conceptual inconsistencies suggests promising applications, which require reasoning under noise or uncertainty.

\begin{figure}[H]
     \centering
     \includegraphics[width=\textwidth]{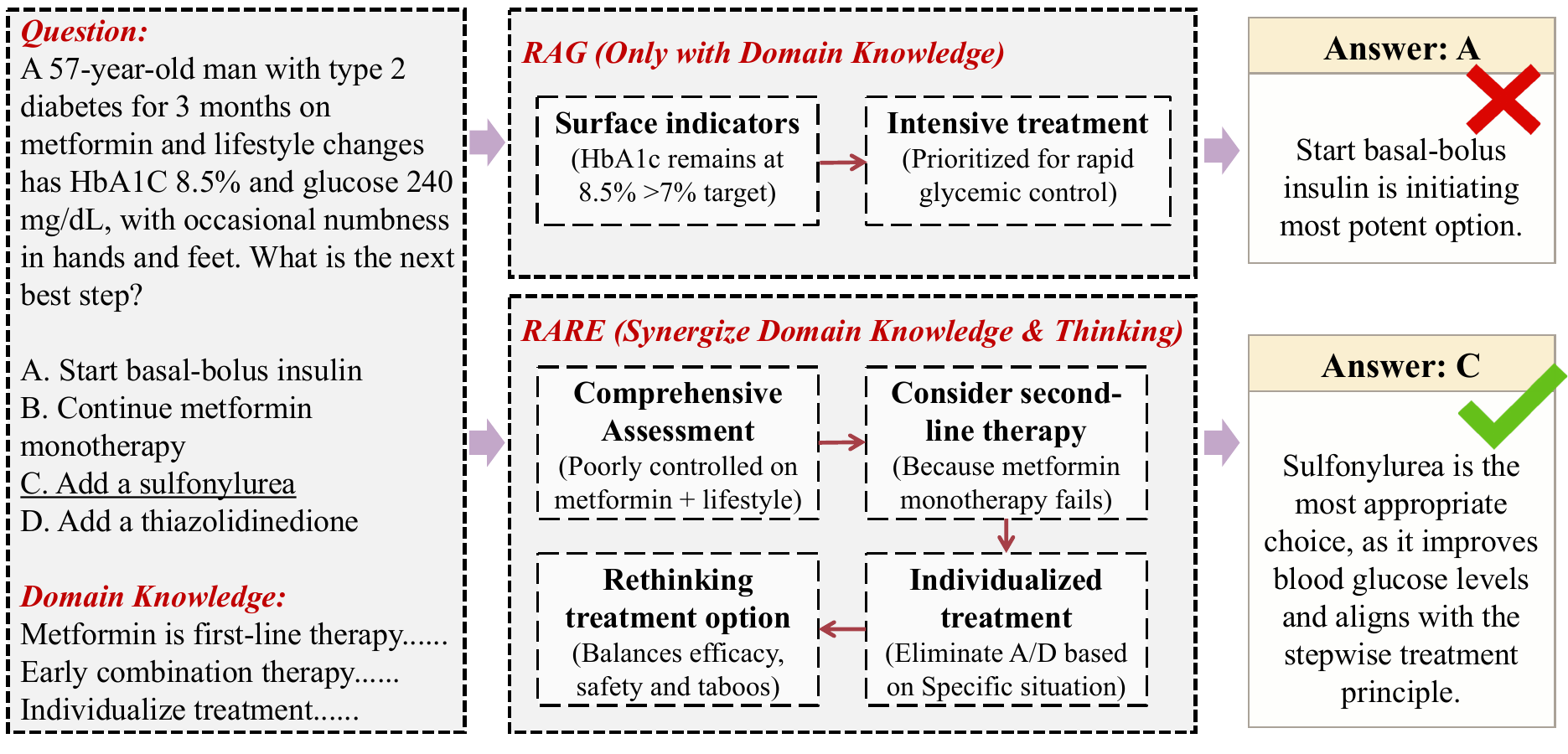}
     \caption{One more case study. Compared to RAG (only with domain knowledge), RARE (combining domain knowledge and thinking) enables LLMs to reason more deeply and accurately. RAG depends only on surface indicators, hastily concluding that the patient requires immediate glucose-lowering intervention, leading to an incorrect answer. In contrast, RARE integrates both clinical indicators and the effectiveness of prior treatment, carefully reasoning that the patient needs second-line therapy while providing a individualized treatment plan—ultimately arriving at the correct answer.}
    \label{fig:Comparsion between RAG and RARE}
\end{figure}

\newpage
\section{Implementation Details}
\label{appendix:implementation}
\subsection{The Setting of Preliminary Experiment}
In our preliminary experiment, we employed the Llama3.1-8B-Instruct model to investigate the effectiveness of \textsc{RARE} in reducing knowledge loss and enhancing reasoning loss by varying the proportion of document content used. Specifically, we utilized the \texttt{en\_core\_web\_sm} model from the \textsc{spaCy} framework to perform named entity recognition on the retrieved full document $R(x)$, extracting, for instance, the top-$k$ most frequent entities (with $k = 100$ in our setting).

The model input was constructed as follows:
\[
\text{Input} = \text{Question}~x + [R(x)]_{\text{part}} + \text{CoT Answer}~ y,
\]
where the CoT answer $y$ was generated by QwQ-32B+RAG. Here, $[R(x)]_{\text{part}}$ denotes a partial inclusion of the retrieved document $R(x)$, such as retaining 25\%, 50\%, or 75\% of its content.

To compute the knowledge loss, we calculated the average token-level loss for the extracted entity tokens that appeared in the CoT answer $y$. The reasoning loss was defined as the average loss of all other tokens in the same segments (excluding the entity tokens).

Finally, we acknowledge that there is a certain gap between knowledge items and named entities, but this substitution is permissible for the preliminary experiment. The more rigorous verification of RARE's effectiveness comes from our extensive main experiments and discussions.

\subsection{The Setting of Main Experiments}
\label{appendix:main experiment}

In this part, we provide a detailed description of the experimental configurations presented in Table~\ref{table:performance of RARE}. For each small model architecture, we evaluated the performance of the base model under six different methodologies: Chain-of-Thought (CoT), Supervised Fine-Tuning (SFT), Retrieval-Augmented Generation (RAG), Retrieval-Augmented Fine-Tuning (RAFT), SFT+RAG, and our proposed RARE method.

The CoT approach requires no training of the base model, relying solely on specific prompts to activate the model's inherent problem-solving capabilities. The SFT method trains the base model using constructed question-answer pairs, enabling direct output of answers. This approach tends to make models memorize training set knowledge rather than develop domain reasoning abilities, resulting in poor performance on test sets due to insufficient domain knowledge and reasoning capabilities. Similarly, RAG is a training-free method that combines retrieved relevant content with test questions into prompts for evaluation. Though RAG uses identical test prompts to SFT+RAG and RARE, and provides external domain knowledge during testing, the lack of inherent reasoning abilities leads to suboptimal performance despite showing significant improvement over CoT.

RAFT represents a closely related work to our research. However, due to its rigid structured output requirements and selection of teacher models lacking reasoning capabilities, it fails to effectively facilitate domain thinking in models, potentially leading to the \textbf{Texas sharpshooter fallacy}. SFT+RAG serves as the most critical baseline comparison for RARE. We selected identical teacher models to those used in RARE for knowledge distillation, trained models with reasoning capabilities, and then employed enhanced RAG prompts for testing. This comparison validates our hypothesis that cultivating domain thinking necessitates the organic integration of Knowledge Integration and Contextualized Reasoning throughout the training process.

Finally, our RARE technique first distills teacher models their ability to solve problems through contextualized reasoning with external knowledge, then uses this data to train smaller models. This process enables small models to acquire domain thinking capabilities, achieving performance on test sets that even surpasses that of the teacher models.

We implemented a series of rigorous and robust experimental configurations to ensure that our results accurately reflect the superiority of our method while guaranteeing reproducibility. In terms of hardware requirements, full experiments need at least 8 GPUs with no less than 64GB of memory.

Similar parameter configurations were adopted for both the distillation teacher model and the trained smaller inference model. The inference process was implemented based on vllm version 0.6.2, with the maximum input sequence length set to 22k tokens and the maximum output length set to 10k tokens. To optimize the balance between output quality and diversity, we configured the temperature at 0.95, top-p at 0.7, and top-k at 50.

For the training phase, we applied Fully Sharded Data Parallel (FSDP) and auto-wrapper techniques to enable distributed training across 8 GPUs. Across all datasets, we consistently conducted training for 5 epochs using a cosine learning rate scheduler. The initial learning rate was set to 1e-5 with a warm-up ratio of 5\%, and a batch size of 64. For better computational efficiency with numerical stability, we implemented bfloat16 mixed-precision. For optimization, we employed the AdamW optimizer with hyperparameters $\beta_1=0.9$, $\beta_2=0.95$, and a weight decay coefficient of 1e-4.

\subsection{The Setting of Multi-Task and Cross-Task Learning}
\label{appendix:multi}
We selected five medical-domain datasets—MedQA, PubMedQA, PubHealth, CoVERT, and BioASQ—and used the correctly distilled outputs (based on QwQ-32B+RAG, detailed discussion
in Appendix~\ref{appendix:Rejection Sampling}) from eight sampling iterations as the complete training set, combining all five datasets. Training was conducted independently on three backbone models: Llama3.1-8B-Instruct, Qwen-2.5-7B-Instruct, and Mistral-7B-Instruct-v0.3. The overall training configuration remained consistent with our main experiments.

For evaluation, each trained model was tested on the same five datasets (as shown in Table~\ref{table:multi_task}). The results show that models trained under the multi-task setting achieve comparable or even superior performance compared to those trained separately on single tasks, providing preliminary evidence for the generalization capability of the RARE training approach.

To further evaluate the generalization of domain-specific reasoning acquired through RARE training, we conducted additional testing on the HLE (Bio/Med) dataset. Specifically, we directly evaluated the checkpoints obtained from the aforementioned multi-task training on the HLE dataset. This evaluation is referred to as Cross-Task testing, as the RARE-trained models had never been exposed to the HLE task during training. For the RAG-based setting, we retrieved relevant documents from MedOmniKB. As shown in Table~\ref{table:multi_task} and Table~\ref{table:hle}, the RARE-trained models demonstrate strong generalization capabilities, even on par with the performance of GPT-4o or DeepSeek-R1+RAG.

\subsection{The Setting of Multi-Modal Experiment}
For multi-modal setting, we changed our teacher model since QwQ-32B does not support multimodal inputs. Instead, we employed Qwen2.5-VL-32B-Instruct to distill the training data required for RARE with contextualized reasoning. Unlike pure text-based tasks, each example in the multi-modal setting typically includes both a textual question and one or more associated images.

We used the entire dataset generated from \textit{a single round of distillation} to train RARE, employing Qwen2.5-VL-7B-Instruct as the student model. The performance is reported in Table~\ref{table:multi-modal}. 

Considering that VQA-RAD is also a medical-domain dataset, we further conducted a multi-task RARE training experiment—following a setup similar to Appendix~\ref{appendix:multi}—using the correctly distilled samples from five medical-domain datasets (MedQA, PubMedQA, PubHealth, CoVERT, and BioASQ, distilled via QwQ-32B+RAG) along with these from VQA-RAD (via Qwen2.5-VL-32B-Instruct+RAG). The trained model achieved an accuracy of 61.4\%, exceeding RAG by 9.2\%.

These results further validate that the RARE paradigm is effective not only for text-only tasks but also extends naturally to multimodal tasks. Moreover, the strong performance in the multi-task setting provides additional evidence for the generalizability of domain reasoning acquired through RARE.

\subsection{The Setting of Parameter-Efficient Fine-Tuning}
We designed three sets of LoRA experiments with rank values set to 32, 64, and 128 respectively. Following conventional parameter optimization strategies, we configured the LoRA alpha values to be twice the corresponding rank values, resulting in alpha values of 64, 128, and 256.

For each experimental group, we inserted LoRA adapters into all linear layers of the model, implementing comprehensive Parameter-Efficient Fine-Tuning (PEFT). To enhance training effectiveness, we employed the advanced LoRA+ technique with a learning rate ratio of 16, allowing LoRA parameters to learn at a rate 16 times faster than regular parameters. Additionally, we set the LoRA dropout rate to 0.1 to ensure sufficient generalization capability of the model. Beyond specific parameters, other configurations for LoRA experiments remained consistent with the main experimental setup.

Through comparative analysis, we found that a rank value of 64 achieved the optimal balance between computational resource consumption and performance. This investigation successfully explored the feasibility and effectiveness of applying parameter-efficient fine-tuning methods within the RARE framework under constrained scenarios, e.g. limited parameter and computation resources.

\subsection{The Setting of Reinforcement Learning}
To begin, we construct the data required for RL training. We adopt the data construction method for Kahneman-Tversky Optimization (KTO), which is relatively straightforward: it involves appending a kto-tag to each response generated by QwQ-32B+RAG (see Appendix~\ref{appendix:Rejection Sampling} for distillation details), indicating whether the selected option is correct relative to the ground truth.

Next, we describe the three RARE training strategies evaluated in Table~\ref{table:rl_part}. RARE (SFT) corresponds to the supervised fine-tuning setup using the same data as in Table~\ref{table:performance of RARE}, with full training details provided in Appendix~\ref{appendix:main experiment}. RARE (KTO) refers to training from scratch using KTO on the base model—Llama3.1-8B-Instruct. In contrast, RARE (KTO based on SFT) initializes from the RARE (SFT) checkpoints and continues training using KTO.

For training hyperparameters, we follow standard configurations (e.g., epoch=3, learning rate=5e-6). However, due to the well-known sensitivity of RL algorithms to hyperparameters—especially on small datasets and models~\citep{paine2020hyperparameter,eimer2023hyperparameters}—we make minor hyperparameter adjustments across different datasets as needed. For example, we increased the number of training epochs to 6 for certain small datasets (e.g., CoVERT) for convergence.

\newpage

\section{Instruction Templates}
\label{app:instructions1}
\begin{tcolorbox}[
    colframe = gray,       
    colback = gray!5!white,             
    coltitle = white,                   
    coltext = black,                   
    fonttitle = \bfseries,              
    title = {RARE Distillation Instruction (also for Testing RARE, RAG, and SFT+RAG)},  
    boxrule = 1pt,                      
    arc = 2mm,                        
    width = \linewidth,                
    left = 7pt,                        
    right = 7pt,                        
    top = 5pt,                          
    bottom = 5pt                        
]
\fontsize{8.5pt}{10pt}\selectfont
You are a professional medical expert to answer the \# Question. Please first think step-by-step using the \# Retrieved Documents and then answer the question. Your responses will be used for research purposes only, so please have a definite answer.\\
The format should be like:\\
<think>\\
...\\
</think>\\
<answer>A/B/C/D</answer> (only one option can be chosen)\\

\# Retrieved Documents\\
\{documents\}\\

\# Question\\
\{question\}\\
\end{tcolorbox}

\label{app:instructions2}
\begin{tcolorbox}[
    colframe = gray,       
    colback = gray!5!white,            
    coltitle = white,                   
    coltext = black,                   
    fonttitle = \bfseries,              
    title = {SFT+RAG Distillation Instruction (also for Testing CoT, SFT)}, 
    boxrule = 1pt,                      
    arc = 2mm,                         
    width = \linewidth,                 
    left = 7pt,                        
    right = 7pt,                        
    top = 5pt,                         
    bottom = 5pt                       
]
\fontsize{8.5pt}{10pt}\selectfont
You are a professional medical expert to answer the \# Question. Please first think step-by-step using your own knowledge and then answer the question. Your responses will be used for research purposes only, so please have a definite answer.\\
The format should be like:\\
<think>\\
...\\
</think>\\
<answer>A/B/C/D</answer> (only one option can be chosen)\\

\# Question\\
\{question\}\\
\end{tcolorbox}

\label{app:instructions3}
\begin{tcolorbox}[
    colframe = gray,       
    colback = gray!5!white,             
    coltitle = white,                   
    coltext = black,                    
    fonttitle = \bfseries,              
    title = {RAFT Distillation Instruction (also for Testing RAFT)},  
    boxrule = 1pt,                      
    arc = 2mm,                          
    width = \linewidth,                 
    left = 7pt,                        
    right = 7pt,                       
    top = 5pt,                          
    bottom = 5pt                        
]
\fontsize{8.5pt}{10pt}\selectfont
\# Question \\
\{question\}\\

\# Retrieved Documents \\
\{documents\}\\

You should Answer \# Question STRICTLY in this FORMAT:\\
\#\#\#Reason:  \\
Use several quotes from \# Retrieved Documents:  \\
- \#\#begin\_quote\#\# [Relevant text 1] \#\#end\_quote\#\#  \\
- \#\#begin\_quote\#\# [Relevant text 2] \#\#end\_quote\#\# \\
Then think step-by-step.   \\
\#\#\#<answer>A/B/C/D</answer>\\
\end{tcolorbox}

\label{app:instructions4}
\begin{tcolorbox}[
    colframe = gray,       
    colback = gray!5!white,            
    coltitle = white,                  
    coltext = black,                    
    fonttitle = \bfseries,             
    title = {RARE (multi-modal) Distillation Instruction (also for Testing RARE, RAG) }, 
    boxrule = 1pt,                     
    arc = 2mm,                         
    width = \linewidth,                 
    left = 7pt,                         
    right = 7pt,                        
    top = 5pt,                         
    bottom = 5pt                        
]
\fontsize{8.5pt}{10pt}\selectfont
You are a professional medical expert in fact-checking, skilled in analyzing the accuracy of \# Statement. Please first think step-by-step using the \# Retrieved Documents and \# Image related and then check \# Statement by using your own knowledge. Your responses will be used for research purposes only, so please have a definite answer. You should respond in the format:\\
<think>\\
...\\
</think>\\
<answer>A/B/C</answer> (only one option can be chosen)\\

\# Retrieved Documents\\
\{documents\}\\

\# Image\\
<image>\\

\# Statement\\
\{statement\}\\
\end{tcolorbox}

\newpage
\section{Impact Statements}
\label{appendix:impact}
We discusses both potential positive and negative societal impacts of the introduced RARE framework. On the positive side, RARE could significantly advance the development of domain-specific intelligence by enabling more efficient and effective reasoning in large language models, potentially leading to breakthroughs in many fields. It may also democratize access to expert-level knowledge and reasoning capabilities, empowering professionals and organizations with more accurate and reliable decision-making tools. However, we also acknowledges possible negative impacts, such as the potential for misuse of the technology in generating misleading or false information if the retrieved knowledge is not properly verified. Additionally, there could be job displacement in certain professions where domain-specific reasoning was previously a human expertise advantage. We suggest that responsible deployment and continuous monitoring of the technology are necessary to mitigate these risks.

\newpage
\section{Cognitive \& Educational Background}
\label{appendix:background}
\subsection{Bloom’s Taxonomy}
Bloom’s taxonomy categorizes cognitive processes into six progressively complex levels (as shown in Fig.~\ref{fig:Overview}): remembering, understanding, applying, analyzing, evaluating, and creating. The first two levels—remembering and understanding—are considered foundational stages of knowledge acquisition, while the latter four represent higher-order cognitive skills~\citep{krathwohl2002revision,bloom1956taxonomy,krathwohl1964taxonomy}. As a seminal framework in educational psychology, Bloom’s taxonomy continues to play a vital role in instructional design, curriculum development, and educational assessment~\citep{armstrong2010bloom}.

\subsection{Domain Thinking \& Polanyi Paradox}
Psychological research has shown that creative thinking possesses both domain-general and domain-specific characteristics~\citep{hong2010creative,toplak2002domain}. Domain-specific thinking refers to paradigmatic cognitive patterns developed within a particular professional context (e.g., the economic supply-demand logic chain illustrated in Fig.~\ref{fig:Overview}), essentially representing structured applications of domain knowledge. This naturally raises a critical question: \textit{if domain thinking is so crucial, why not directly teach models to learn these paradigms?}

The philosopher Michael Polanyi offers a classic explanation through his cognitive paradox: explicit knowing—what can be codified and articulated—constitutes only the tip of the iceberg, whereas a vast amount of tacit knowing remains largely inexpressible through language~\citep{polanyi1962tacit,gourlay2002tacit}. As demonstrated in the complex diagnostic logic outlined in Table~\ref{tab:case-study-part1} and Table~\ref{tab:case-study-part2}, such tacit knowing is difficult to formalize but crucial for reasoning. While domain thinking cannot be directly taught, it can be progressively acquired through case-based learning~\citep{autor2014polanyi}. This underpins the core rationale for referring to the reasoning process in this study as \textit{contextualized reasoning}.

Cognitive psychology further reveals that individuals with advanced cognitive skills exhibit strong context-adaptive regulation~\citep{sa1999domain}. Large language models (LLMs) such as QwQ-32B and DeepSeek-R1 have demonstrated domain adaptability on par with human experts: with minimal prompting (see Appendix~\ref{app:instructions1}), they are capable of specialized reasoning. This level of cognitive flexibility—mirroring that of human experts—is a key reason for selecting QwQ-32B as the teacher model in this study.

%% file: Sections/Check_List.tex
\newpage
\section*{NeurIPS Paper Checklist}

\begin{enumerate}

\item {\bf Claims}
    \item[] Question: Do the main claims made in the abstract and introduction accurately reflect the paper's contributions and scope?
    \item[] Answer: \answerYes{} %
    \item[] Justification: The main claims are clearly presented in the abstract and introduction, and are further elaborated and substantiated throughout the subsequent sections of the paper.

\item {\bf Limitations}
    \item[] Question: Does the paper discuss the limitations of the work performed by the authors?
    \item[] Answer: \answerYes{} %
    \item[] Justification: Please refer to the Section \ref{sec:conclusion} in the paper.

\item {\bf Theory assumptions and proofs}
    \item[] Question: For each theoretical result, does the paper provide the full set of assumptions and a complete (and correct) proof?
    \item[] Answer: \answerYes{} %
    \item[] Justification: Please refer to the Figure \ref{fig:preliminary experiment} and Appendix \ref{appendix:implementation} in the paper.

    \item {\bf Experimental result reproducibility}
    \item[] Question: Does the paper fully disclose all the information needed to reproduce the main experimental results of the paper to the extent that it affects the main claims and/or conclusions of the paper (regardless of whether the code and data are provided or not)?
    \item[] Answer: \answerYes{} %
    \item[] Justification: Please refer to the Section \ref{sec:method} and Appendix \ref{appendix:implementation} in the paper.

\item {\bf Open access to data and code}
    \item[] Question: Does the paper provide open access to the data and code, with sufficient instructions to faithfully reproduce the main experimental results, as described in supplemental material?
    \item[] Answer: \answerYes{} %
    \item[] Justification: We provide well-documented code repositories in the supplementary materials. Additionally, we commit to making the code publicly available by the time of the camera-ready submission.

\item {\bf Experimental setting/details}
    \item[] Question: Does the paper specify all the training and test details (e.g., data splits, hyperparameters, how they were chosen, type of optimizer, etc.) necessary to understand the results?
    \item[] Answer: \answerYes{} %
    \item[] Justification: Please refer to the Section \ref{sec:experiments} and Appendix \ref{appendix:implementation} in the paper.

\item {\bf Experiment statistical significance}
    \item[] Question: Does the paper report error bars suitably and correctly defined or other appropriate information about the statistical significance of the experiments?
    \item[] Answer: \answerYes{} %
    \item[] Justification: Please refer to the Section \ref{sec:experiments} and Appendix \ref{appendix:implementation} in the paper.

\item {\bf Experiments compute resources}
    \item[] Question: For each experiment, does the paper provide sufficient information on the computer resources (type of compute workers, memory, time of execution) needed to reproduce the experiments?
    \item[] Answer: \answerYes{} %
    \item[] Justification: Please refer to the Section \ref{sec:experiments} and Appendix \ref{appendix:implementation} in the paper.
    
\item {\bf Code of ethics}
    \item[] Question: Does the research conducted in the paper conform, in every respect, with the NeurIPS Code of Ethics \url{https://neurips.cc/public/EthicsGuidelines}?
    \item[] Answer: \answerYes{} %
    \item[] Justification: This paper conforms, in every respect, with the NeurIPS Code of Ethics.

\item {\bf Broader impacts}
    \item[] Question: Does the paper discuss both potential positive societal impacts and negative societal impacts of the work performed?
    \item[] Answer: \answerYes{} %
    \item[] Justification: Please refer to the Appendix \ref{appendix:impact} in the paper.
    
\item {\bf Safeguards}
    \item[] Question: Does the paper describe safeguards that have been put in place for responsible release of data or models that have a high risk for misuse (e.g., pretrained language models, image generators, or scraped datasets)?
    \item[] Answer: \answerNA{} %
    \item[] Justification: The paper poses no such risks.

\item {\bf Licenses for existing assets}
    \item[] Question: Are the creators or original owners of assets (e.g., code, data, models), used in the paper, properly credited and are the license and terms of use explicitly mentioned and properly respected?
    \item[] Answer: \answerYes{} %
    \item[] Justification:  The creators and original owners of all assets used in the paper are properly credited, and the licenses and terms of use are explicitly mentioned and fully respected.

\item {\bf New assets}
    \item[] Question: Are new assets introduced in the paper well documented and is the documentation provided alongside the assets?
    \item[] Answer: \answerNA{} %
    \item[] Justification: The paper does not release new assets.

\item {\bf Crowdsourcing and research with human subjects}
    \item[] Question: For crowdsourcing experiments and research with human subjects, does the paper include the full text of instructions given to participants and screenshots, if applicable, as well as details about compensation (if any)? 
    \item[] Answer: \answerNA{} %
    \item[] Justification: The paper does not involve crowdsourcing nor research with human subjects.

\item {\bf Institutional review board (IRB) approvals or equivalent for research with human subjects}
    \item[] Question: Does the paper describe potential risks incurred by study participants, whether such risks were disclosed to the subjects, and whether Institutional Review Board (IRB) approvals (or an equivalent approval/review based on the requirements of your country or institution) were obtained?
    \item[] Answer: \answerNA{} %
    \item[] Justification:  The paper does not involve crowdsourcing nor research with human subjects.

\item {\bf Declaration of LLM usage}
    \item[] Question: Does the paper describe the usage of LLMs if it is an important, original, or non-standard component of the core methods in this research? Note that if the LLM is used only for writing, editing, or formatting purposes and does not impact the core methodology, scientific rigorousness, or originality of the research, declaration is not required.
    \item[] Answer: \answerYes{} %
    \item[] Justification: Please refer to the Section \ref{sec:experiments} and Appendix \ref{appendix:implementation} in the paper.

\end{enumerate}